\documentclass[3p]{elsarticle}

\usepackage[dvipsnames]{xcolor}
\usepackage{lettrine}%  dropped capital letter at beginning of paragraph
\usepackage{graphicx}
\usepackage{xcolor}
\usepackage{enumitem}
\usepackage{amsmath}
\usepackage{subfigure}
\usepackage[subfigure]{tocloft}
\usepackage{amsfonts}
\usepackage{listings}
\usepackage{color}
\usepackage{multirow}
\usepackage{amssymb}
\usepackage{cprotect} % Allows \verb within command arguments, like \caption{}. See https://tex.stackexchange.com/a/8814
\definecolor{codegreen}{rgb}{0,0.6,0}
\usepackage[T1]{fontenc}
\usepackage[scaled=0.85]{beramono}
\makeatletter
\newcommand\monottfamily{%
  \def\fvm@Scale{0.75}% scales the font down
  \fontfamily{fvm}\selectfont% selects the Bera Mono font
}
\makeatother

\lstdefinestyle{commonstyle}{
literate={-}{-}1
}
\lstdefinelanguage{csp}{
  keywords=[1]{assertion, untimed, csp, refines, timed, is, deadlock, free, reachable, associated},
  keywords=[2]{nametype,channel,Timed,if,then,else, timed_priority,WAIT,SKIP,STOP,USTOP,!,?,[],->,[],::,[|,|],\{|,|\}},
  sensitive=true, %keywords are not case sensitive
  breaklines=true, % wrap lines if they don't fit
  comment=[l]{//},
  morecomment=[s]{/*}{*/},
  morecomment=[l]{--},
  commentstyle=\color{codegreen},%\ttfamily
  moredelim=[is][\color{Plum}\bfseries]{/@}{@/},
}

\newcommand{\RC}[1]{{\sf #1}}
\usepackage{url}

\usepackage{breakurl}
\usepackage[breaklinks]{hyperref}
\usepackage[capitalise]{cleveref}
\crefname{issue}{issue}{issues}
\crefname{property}{property}{properties}
\crefname{lstfloat}{Listing}{Listings}
\creflabelformat{issue}{#2(#1)#3}
\crefrangelabelformat{issue}{#3#1#4-#5#2#6}
\crefname{line}{line}{lines}
\crefrangelabelformat{line}{#3#1#4-#5#2#6}
\newcommand{\CSPM}[0]{\texttt{CSP$_\text{\texttt{M}}$}}
\usepackage{float}
\newfloat{lstfloat}{htbp}{lop}
\floatname{lstfloat}{Listing}

\hypersetup{
  colorlinks,
  linkcolor={blue!50!black},
  citecolor={blue!50!black},
  urlcolor={blue!80!black}
}

\usepackage[skins]{tcolorbox}
\usepackage{fmtcount}
\usepackage{imakeidx}
\makeindex[title=Index of Changes, name=changes]

\ifdefined\CHANGES



\newlistof{changes}{s}{\listchangename}
\newcommand{\changed}[2][?]{%
  \refstepcounter{changes}%
  \linkdest{\thechanges}%
  {\color{blue}#2}\textsuperscript{#1}%
  \addcontentsline{s}{changes}{\protect\numberline{\thechanges}#1}%
}%

\newtcolorbox{echanged}[1]{enhanced,title=#1, sharp corners,colback=blue!20,
  attach boxed title to top right=
  {xshift=0mm,yshift=0mm},
  boxed title style={size=small,colback=blue},
  size=minimal,
  finish={
    \refstepcounter{changes}
    \addcontentsline{s}{changes}
    {\protect\numberline{\thechanges}#1}
  }
}
\fi
\newcommand{\notinpaper}[1]{%
  \index[changes]{Comment \textbf{C\expandafter\sortgref#1\sortgref} not reflected in the text.\removecomma|HIDE}%
}

\newrobustcmd{\removecomma}[1]{}
\newcommand{\HIDE}[1]{}

\newcommand{\C}[1]{%
  \index[changes]{Comment \textbf{C\expandafter\sortgref#1\sortgref}|BH{\arabic{changes}}}{C#1}%
}%

\def\sortgref#1\sortgref{%
  \ignoresort{\ifnum#1<10 00\else\ifnum #1<100 0\fi\fi#1}#1%
}
\protected\def\ignoresort#1{}
\ifdefined\CHANGES

\else
\newcommand{\listchangename}{List of Changes}
  \newlistof{changes}{s}{\listchangename}
  \newcommand{\changed}[2][?]{#2}
  
\newtcolorbox{echanged}[1]{size=minimal}
\fi

\begin{document}
\begin{frontmatter}
\title{Safety assurance of an industrial robotic control system using hardware/software co-verification} 
\author[inst1]{Yvonne Murray\corref{cor1}}
\ead{yvonne.murray@uia.no}
\author[inst1]{Martin Sirevåg}
\ead{martin.sirevag@uia.no}
\author[inst2]{Pedro Ribeiro}
\ead{pedro.ribeiro@york.ac.uk}
\author[inst1,inst3]{David A. Anisi}
\ead{david.anisi@nmbu.no}
\author[inst4]{Morten Mossige}
\ead{morten.mossige@no.abb.com}

\address[inst1]{Dept.\ of Mechatronics, Faculty of Engineering and 
Science, University of Agder (UiA), Norway}
\address[inst2]{Dept.\ of Computer Science, University of York, UK}
\address[inst3]{Robotics Group, Faculty of Science \& Technology, Norwegian University of Life Sciences (NMBU), Norway}

\address[inst4]{ABB Robotics, Bryne, Norway}

%\maketitle              % typeset the header of the contribution
%
\begin{abstract}
As a general trend in industrial robotics, an increasing number of safety functions are being developed or re-engineered to be handled in software rather than by physical hardware such as safety relays or interlock circuits. This trend reinforces the importance of supplementing traditional, input-based testing and quality procedures which are widely used in industry today, with formal verification and model-checking methods. To this end, this paper focuses on a representative safety-critical system in an ABB industrial paint robot, namely the High-Voltage electrostatic Control system (HVC). The practical convergence of the high-voltage produced by the HVC, essential for safe operation, is formally verified using a novel and general co-verification framework where hardware and software models are related via platform mappings. This approach enables the pragmatic combination of highly diverse and specialised tools. The paper's main contribution includes details on how hardware abstraction and verification results can be transferred between tools in order to verify system-level safety properties. It is noteworthy that the HVC application considered in this paper has a rather generic form of a feedback controller. Hence, the  co-verification framework and experiences reported here are also highly relevant for any cyber-physical system tracking a setpoint reference.

\end{abstract}
\end{frontmatter}
\section{Introduction}
\label{sec:intro}
\lettrine[nindent=0pt, lines = 2]{T}{}he liberation of industrial robots from traditional metal cages and steadily increasing number of co-bots working side by side with humans are illustrative examples of a general trend in industrial robotics. In the wake of this, more and more safety-critical functions are now being developed to be handled by software and/or firmware components instead of hardware safety relays or interlock circuits. Modern industrial robots are heavily dependent on software-implemented safety signals to monitor and control various critical subsystems such as current/voltage supervision and emergency stop or short circuit interrupts. This trend brings several distinctive advantages such as cost-reduction and increased flexibility. Nevertheless, it also introduces or reinforces negative side-effects, most notably in the form of higher system complexity, vulnerability and dependability~\cite{safety_critical_robots_survey_2017}. 

To set the stage for and address this ongoing industrial trend, this paper advocates use of formal verification techniques, which can provide an extra level of assurance by verifying the logic of a system.
The application of formal methods in the robotics industry will ideally help to identify potential pitfalls at a much earlier phase of the development cycle~\cite{10.5555/2930832} and serve as an important supplement to the traditional testing and safety risk identification and mitigation actions which are already employed~\cite{model_checking_industrial_robot_systems_2011}. Obtaining sufficiently high testing coverage in complex industrial systems can be time-consuming and expensive. In practice, it is most often not viable to account for every scenario, which means that testing can fail to reveal potential safety-critical issues.

The HVC system considered in this paper provides a perfect example of this. As described in~\cite{FV_of_HVC_SBMF_2020}, a previous version of the HVC software (SW) has been shown to contain some errors, e.g., failure to properly follow the given setpoint. These errors are described in more detail in~\cite{FV_of_HVC_SBMF_2020} and went undetected despite passing rigorous and certified quality assurance and testing procedures. These included \emph{a priori} and systematic identification of risk mitigation plans (e.g., using HAZID/HAZOP), as well as thorough testing procedures consisting of static code analysis, unit testing, component testing and system test~I and~II. Here, system test~I encompasses hardware tests with the Integrated Painting System (IPS) and HVC active, while system test~II entails testing of the entire robotic system using actual paint.

The robotic spray booth in, e.g., a car factory, may contain flammable solvent and paint particles in the air. Hence, paint robots are certified for operation in potentially explosive atmospheres in accordance with regional ATEX/NFPA/IECEx standards (ATEX Directive – 2014/34/EU, IEC 60079). The paint version of the ABB Industrial Robot Controller unit, denoted IRCP, is certified with respect to the ISO~10218 standard for safety requirements for industrial robots. Paint robots using HVC are also certified according to the EN50176 standard for using high-voltage in explosive environments, while the paint atomizer is certified in accordance to ISO~9001 and ISO~14001.

\begin{figure}[t]
\centering\includegraphics[width=0.8\textwidth]{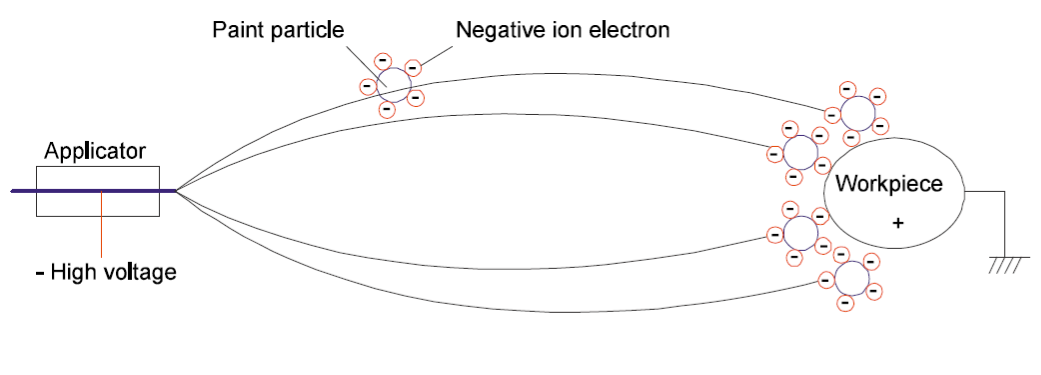}
\caption{In electrostatic painting, high-voltage (approximately 40-90 kV) charges the paint particles at
the applicator. The particles follow the lines of the electrostatic field from the applicator (cathode)
to the earthed workpiece (anode).}\label{fig:painting}
\end{figure}

Industrial paint robots use high-voltage to perform electrostatic painting, where particles are electrically charged and attracted to the grounded paint object, as seen in~\cref{fig:painting}~\cite{thesis:morten,thesis:nina}. In this way, painting quality is ensured while paint consumption and costs are minimized. However, the use of high-voltage also poses certain risks of electric shock and ignition. Fire in the painting cell may result in costly production delays, as well as damage to the equipment. Therefore, it is of great importance that the HVC works as intended.

The HVC example illustrates the fact that the complete elimination of all errors is most often not practical (due to cost and/or time constraints) or even possible. Formal verification provides us not only with a mathematically sound formalism for the specification and verification of robotic systems which ensures correctness, but also provides evidence for safety certification purposes. In fact, a survey on safety-critical robot systems~\cite{safety_critical_robots_survey_2017} recognises formal verification and correct-by-construction control synthesis as two main areas needed to develop safe robot systems.

Robot control systems, like the HVC, have rather natural and generic properties that are expected to be fulfilled by any feedback controller tracking a setpoint reference. Formally verifying that overall the setpoint is followed, a system property, however, requires reasoning over the combined, time-dependent behaviour of software and hardware. For pragmatic reasons these are often modelled using diverse languages and formalisms, making holistic reasoning challenging.

Inspired by co-simulation approaches~\cite{GomesTBLV18}, in this paper we propose a novel and generic co-verification approach for pragmatic verification of system properties. 
\changed[\C{1}]{Models of the software and hardware are coupled through platform mappings that define how the inputs and outputs of the software are connected to those of the hardware, in terms of its sensors and actuators.} With our approach, behavioural properties of individual models -- that may be established using separate domain-specific tools -- can be combined to support the verification of system properties, using practical techniques, such as model checking~\cite{BaierChristel2008Pomc}.

To illustrate the use of co-verification in a representative industrial case study, the HVC software is modelled in RoboChart~\cite{RoboStarTechnology2021,RoboChartSoSym,RoboChart}, while the hardware is modelled in Simulink~\cite{Simulink}. RoboChart is a domain-specific language for model-based software engineering of robotics, with a formal semantics encompassing timed and functional aspects, that is tailored for formal verification. Simulink, on the other hand, is a \emph{de facto} standard for control engineering, as typically used in industry for dynamic simulation. For co-verification we use the MathWorks Simulink Design Verifier (SDV) toolbox~\cite{Simulink}, and the CSP model-checker FDR~\cite{FDR}, as integrated into RoboTool~\cite{RoboStarTechnology2021,RoboChartSoSym,RoboChart}.

Importantly, we demonstrate the value of our approach in identifying errors that existed in an early-phase HVC software  version as described in~\cite{FV_of_HVC_SBMF_2020}. In the next phase, once the identified software shortcomings had been rectified, we were able to show that it satisfies all safety properties of concern. Namely, that overall the system tracks the high-voltage setpoint as set by an operator, and that the software resets the setpoint and disables the high-voltage if it senses \changed[\C{2}]{that} the power supply is unstable. This serves as a testimony of the strength and suitability of using formal verification methods for industrial safety-critical systems.

Some initial and preliminary results of our work regarding formal verification of HVC of industrial paint robot have been previously published in~\cite{FV_of_HVC_SBMF_2020}. This paper extends that work by addressing some fundamental and important aspects, most notably by:
\begin{enumerate}
    \item taking into account the timed aspects of the HVC controller using the timed semantics of RoboChart.
    \item replacing the simplified, binary representation of the output voltage \emph{following} the setpoint, with a real representation and considering timed and dynamic \emph{convergence} towards the setpoint signal.   
    \item providing a crisp dichotomy between control software and physical hardware  parts of the HVC system, together with detailed platform mapping in-between\cite{PMODELR_TechReport,diagrammatic_RoboSim_models}.
    \item modelling the system dynamics of the hardware in Simulink~\cite{thesis:morten,thesis:nina}. 
    \item using a novel approach to co-verification to combine the results from hardware simulations in Simulink with the model-checking capabilities of RoboTool~\cite{RoboTool} to verify that the high-voltage produced by the HVC follows the setpoint, a system-level safety property.
 \end{enumerate}

The remainder of this paper is structured as follows. \cref{sec:related-work} discusses related work. \cref{sec:HVC} provides an overview of the HVC system, contains formulations of the properties to be formally verified (\cref{sec:properties}) and presents a simplified finite state machine of the HVC (\cref{sec:finite}). \cref{sec:co-verification} constitutes the main body of the current paper. It details the co-verification framework and explains how the state machine was modelled in RoboChart and combined with Mathworks SDV simulation and verification results in order to verify the system-level safety property concerning the high-voltage. \cref{sec:cover:software-properties} reports on the verification of software properties. Finally,~\cref{sec:conclusion} provides some discussion and conclusions, as well as suggestions for further research.

\section{Related Work}
\label{sec:related-work}
The survey on safety-critical robot systems in~\cite{safety_critical_robots_survey_2017} identifies seven areas that need further focus and research in order to develop safe, dependable robotic systems. It is notable that at least five of these areas are relevant in the context of this paper, namely: adaptive safety monitoring, modeling and simulation for safety analysis, formal methods for verification, correct-by-construction control, and certification.

A recent survey~\cite{FV_in_robotics_survey_2019} maps and lists the current challenges, used formalisms, tools, approaches, as well as limitations when considering formal specification and verification of autonomous robotic systems. The main results\changed[\C{2}]{} reveal that temporal logic, state-transition and  model checking are the main formalisms and approaches used during the last decade. At the same time, the lack of appropriate tools and sheer resistance to \changed[\C{2}]{adopting} formal verification methods in robotic systems development are recognised as the main limiting factors for wider impact. Likewise, the lack of interoperability and need to capture the essence of complex, industrial robotic systems using several formalisms and tools is recognised. 

Simulation plays an important role in the development of robotic systems, and more widely in the domain of cyber-physical systems (CPS). However, current practice makes it difficult to soundly reason across models of the software, simulation, and hardware, which can exacerbate the reality gap. Co-simulation approaches~\cite{GomesTBLV18,CavalcantiMPW17} bridge the heterogeneity of tools via orchestration, for example, using a common API as advocated in the FMI standard~\cite{FMI}.
Besides the issue of code portability between tools, robotics simulators~\cite{Afzal2000} tend to use different physics engines. A related approach~\cite{diagrammatic_RoboSim_models} to our work on co-verification, extends the diagrammatic simulation language RoboSim \cite{CavalcantiSMRFD19} with facilities to cover physical modelling of robotics and establish formal links between sensors, actuators, and the software, via platform mappings.

Kawahara et al.~\cite{Kawahara2009} address the co-simulation of Simulink and a subset of SysML~\cite{OMG12}, where data is exchanged between models via input/output ports modelled as S-Functions in Simulink. Their focus is on testing of simulations against timed constraints expressed via sequence diagrams using the UML-MARTE~\cite{SG13} profile. Cavalcanti et al.~\cite{CavalcantiMPW17} give semantics to a version of INTO-SysML~\cite{Amalio2015}, a SysML profile suitable for co-simulation using FMI, where RoboChart is used with Simulink in the co-simulation of a chemical detector robot. Bernardeschi et al.~\cite{Bernardeschi2018} use timed automata, encoded for reasoning in the PVS theorem prover~\cite{PVS}, in co-simulation with a Simulink model of a cardiac pacemaker. Our focus, however, is on (co-)verification.

Recognising the broad range of aspects in the engineering of robots, the use of specialised, and complementary, verification techniques is widely reported in the literature. Webster et al.~\cite{Webster2020} propose a ``corroborative'' approach where agreement is sought between different verification techniques with respect to functional requirements, including model checking with PRISM~\cite{PRISM}, simulation-based testing and user validation.

Cardoso et al.~\cite{Cardoso2020} use different methods to verify components of a simulation of NASA's Curiosity rover, where high-level control is driven by a Beliefs-Desires-Intentions (BDI) agent. The agent is verified using the Agent Java Path Finder (AJPF) model checker~\cite{Bordini2008}, while its interface with the environment is verified using Dafny~\cite{Leino2010}. FDR is used to verify a CSP model of the action library nodes that implement control methods following the publish-subscribe paradigm of the Robot Operating System (ROS)~\cite{ROS}. The emphasis is on verification of components with formal models guiding the generation of runtime monitors.

Related, Bourbouh et al.~\cite{Bourbouh2021} report on the combined use of several methods and tools in the development of an assurance case for an inspection rover,
which is modelled in AADL~\cite{FG12}, Simulink and Event-B~\cite{EventB}. Functional requirements are stated using the structured natural language accepted by FRET~\cite{Giannakopoulou2020}, with semantics given in Linear Temporal Logic (LTL) suitable for analysis with Lustre~\cite{LUSTRE} models generated from Simulink via CoCoSim~\cite{Bourbouh2020}, a framework for design, code generation and analysis of discrete dataflow models. Simulink blocks modelling the rover architecture are annotated with assume-guarantee contracts based on component requirements formalised in FRET. System-level properties are then verified via model checking with Kind2~\cite{Kind2}, while some components are verified using Event-B instead.

The literature is rich in approaches for formal verification of Simulink models. Reicherdt and Glesner~\cite{Reicherdt2014} propose translating discrete-time Simulink models into Boogie~\cite{Barnett2005} for verifying the absence of common error classes, such as overflows, underflows, division-by-zero and range violations. CoCoSim follows on from previous work~\cite{Tripakis2005} targeting Lustre and SCADE~\cite{SCADE}. Bostr{\"{o}}m and Wiik~\cite{BostromW16} propose a compositional approach for verifying Simulink blocks annotated with assume-guarantee contracts.

Applications of formal verification methodologies within the control and CPS community have mainly adopted the hybrid system and automata framework of Alur et al.~\cite{FV_hybrid_automata_alur_1993,FV_of_hybrid_systems_Alur_2011}. In this setting, finite- and infinite-time reachability constitute the main verification tools, but unfortunately turn out to be an undecidable problem in general, leaving conservative set approximation as the only viable approach~\cite{reachability_verification_1998,undecidable_reachable_sets_1998}. Hybrid automata also assumes having infinite accuracy and instantaneous reaction which serves as a noticeable discrepancy to the real system and implementation; potentially invalidating the verification results~\cite{open_problems_FV_of_robotic_systems}.

Focusing on formal verification of industrial robot applications, in~\cite{model_checking_industrial_robot_systems_2011}, industrial robot- and PLC-programs are compiled into PROMELA models as input for the SPIN model checker~\cite{SPIN_model_checker_1997}. The work is however restricted to LTL formulas. It further differs from our work by solely considering deadlocks, collisions and kill-switch violations. Narrowing down to industrial paint robots,~\cite{parametric_FV_paint_robot_2017} considers formal verification of paint spraying using the ARIADNE tool for reachability analysis. The focus here is solely on parametric design verification.

\begin{figure}[t]
\centering
\includegraphics[width=0.95\textwidth]{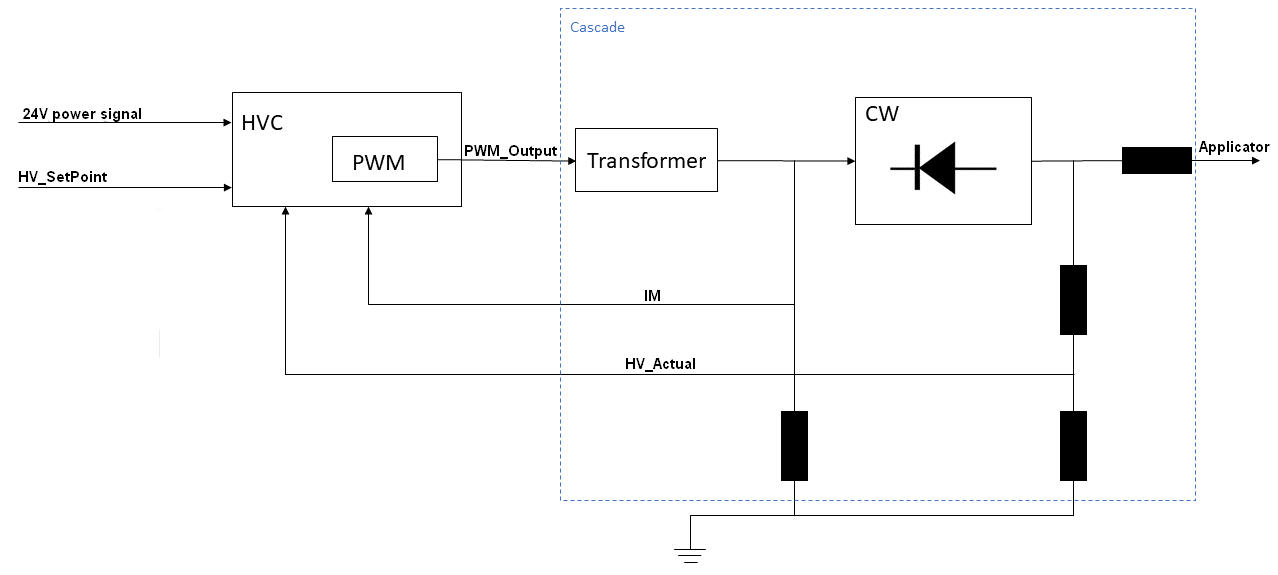}
\caption{Block diagram of one part of the paint robot, containing the HVC.} \label{HVC}
\end{figure}

\section{High-Voltage Control (HVC)}\label{sec:HVC}
\begin{figure}[!htbp]
\centering
\includegraphics[width=0.75\textwidth]{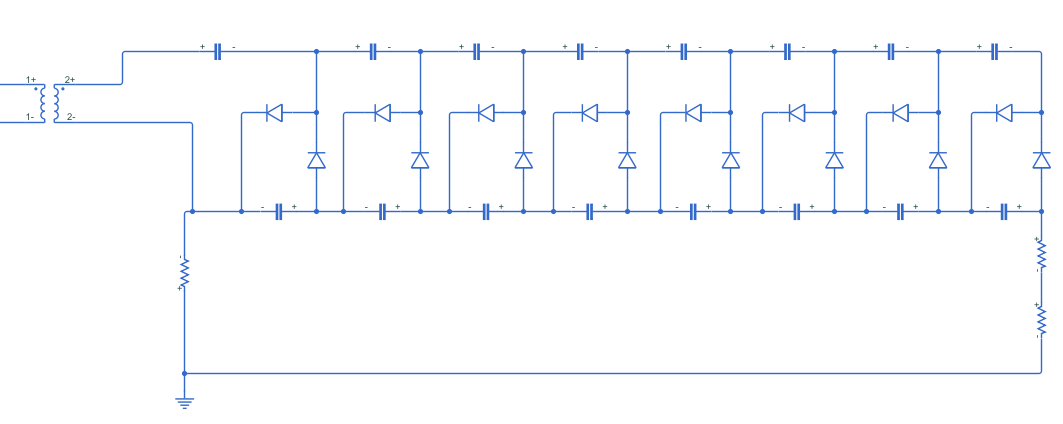}
\caption{The cascade modelled in Simulink with Simscape components, with the transformer, resistors and the Cockroft Walton voltage multiplier circuit.} \label{fig:CW_simscape}
\end{figure}

A simplified block diagram of the HVC part of the paint robot can be seen in~\cref{HVC}.
The HVC module runs the control software loops and associated control logic. Here, the $r(t) = HV\_SetPoint$ signal is a function of time, t, and used as \emph{a priori} given reference for the desired voltage level on the HVC, while the 24V power
signal provides the HVC with electrical power. The $u(t) = PWM\_Output$ signal serves as input signal to the Pulse Width Modulation (PWM) hardware. It is a percentage, from 0 to 100\%, mapped to an analog 0 to 10 voltage signal, which is then increased in the transformer. In the Cockcroft–Walton (CW) cascade generator, there are several voltage doubling circuits, and the voltage is rectified and further increased, before arriving to the applicator, see~\cref{fig:CW_simscape}. Finally, $\bar{y}(t) = [IM; HV\_Actual]^T$ denote current and voltage measurements, respectively, which are fed back into the HVC. It is further noticeable that from a paint robot application point of view, it is given that $HV\_SetPoint \in 0 \cup [30 \: 90],  kV$, that is, once the high-voltage is activated and turned on, it requires values larger than $30  kV$, and that the $r(t) = HV\_SetPoint$ reference value does not change very often, and never faster than within 10 seconds from the previous change. These facts will be used subsequently in order to formally capture and verify some basic properties for HVC. 

\begin{figure}[!htbp]
\centering
\includegraphics[width=0.7\textwidth]{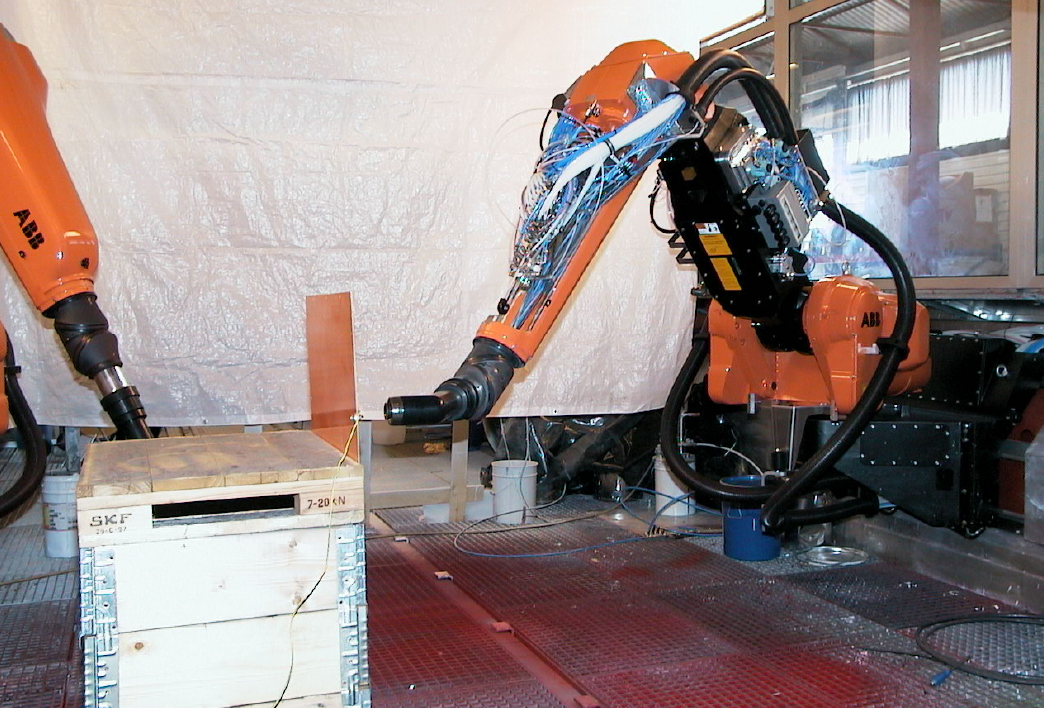}
\caption{Setup for experimental testing with paint using ABB robot. Photo courtesy ABB, from~\cite{thesis:morten}.} \label{fig:lab_setup}
\end{figure}

\begin{figure}[!htbp]
\centering
\includegraphics[width=0.85\textwidth]{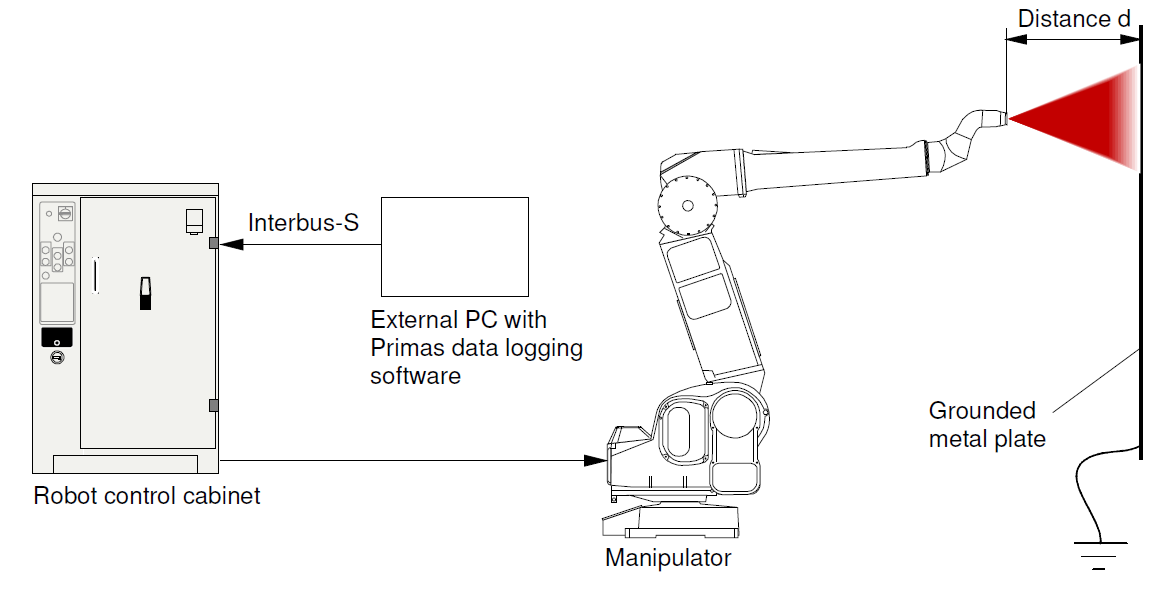}
\caption{Schematic overview of the lab setup for testing ABB paint robot. Photo courtesy ABB, from~\cite{thesis:morten}.} \label{fig:lab_setup_schematic}
\end{figure}

Following the line of thought in~\cite{PMODELR_TechReport,diagrammatic_RoboSim_models}, in order to distinguish and describe both the control software and physical hardware components of the HVC system, a faithful model of the PWM hardware is needed. The PWM hardware comprises the components inside the dashed blue box in~\cref{HVC}, that is, the transformer, CW cascade block and resistors.
\cref{fig:CW_simscape} depicts the diodes, capacitors and resistors defining a CW cascade block as modelled in Simulink Simscape, which allows modelling of physical components and systems. It is noteworthy that by design, each section of the CW block will double the input voltage so that the output voltage of a CW cascade with $N$ sections will equal $2 N V_{in}$. The Simulink models used in this work are based on and extracted from experimental laboratory tests performed in~\cite{thesis:morten,thesis:nina} on real ABB paint robots as depicted in~\cref{fig:lab_setup,fig:lab_setup_schematic}. This serves as a back-drop and starting point for our work.

\begin{figure}[!htbp]
\centering
\includegraphics[width=0.85\textwidth]{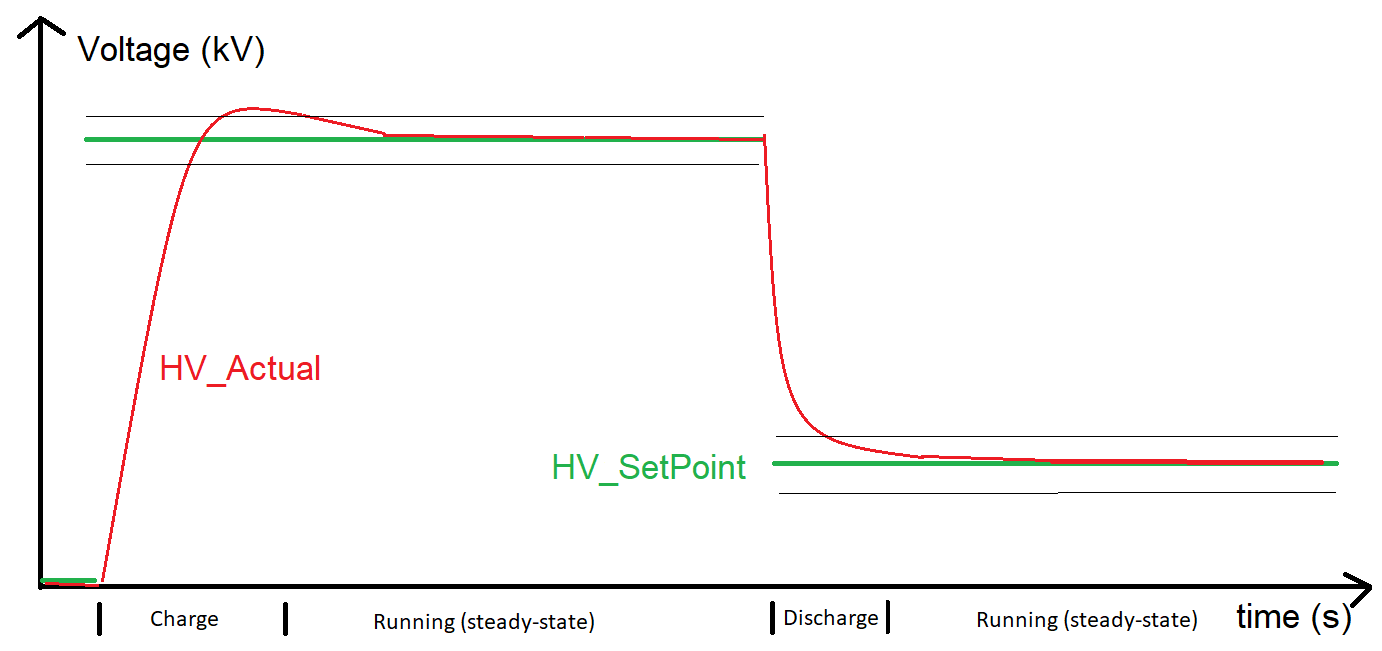}
\caption{The HVC hardware and controller has three distinct \textit{modes}: Charge, Running (steady-state) and Discharge. The Integrated Painting System (IPS) parameters \textit{RampLimit} and \textit{TauPeriod} provide upper \changed[\C{2}]{limits} on the duration of the Charge and Discharge modes respectively.} \label{fig:3_phases_of_PWM_HW}
\end{figure}

The paint robot HVC application has some further distinguishable structure and dynamics that will need to be considered and incorporated into our formal verification scheme. As detailed in~\cite{thesis:morten,thesis:nina}, the PWM hardware model and cascade controller are based on three distinct \textit{modes} as graphically illustrated in~\cref{fig:3_phases_of_PWM_HW}:
\begin{itemize}
    \item \textbf{Charge:} when a new external setpoint, \textit{HV\_SetPoint(t)}, with higher value than the current one arrives and the PWM hardware is ramping up the control signal, $u(t) = PWM\_Output(t)$, in order to increase the value of $HV\_Actual(t)$.
    \item \textbf{Running (steady-state):} when $HV\_Actual$ has converged to $HV\_SetPoint$ and reached a steady state.
    \item \textbf{Discharge:} when the external $HV\_SetPoint$ is set to a lower value and PWM hardware is discharging so that $HV\_Actual$ converges to $HV\_SetPoint$. 
\end{itemize}

It is further noticeable that by design~\cite{ABB_IPS_reference_manual}, there are additional limits on peak deviation between $HV\_Actual$ and $HV\_SetPoint$ as well as the time duration of the Charge and Discharge modes. Namely, a parameter \textit{RampLimit} determines the maximum time in seconds that it will take to ramp up the high-voltage from minimum to maximum level, i.e., from $0$ to $90$ (kV). The default value of \textit{RampLimit} is $2$ seconds. Likewise, it is known that it will take \textit{TauPeriod} seconds for $HV\_Actual$ to reach a level of $30\%$ above a new lower $HV\_SetPoint$ value. The default value of \textit{TauPeriod} is $3$ seconds. Additionally, there are maximum allowed over\changed[\C{2}]{} and under voltage limits. As mentioned earlier, the HVC application, once activated and turned on, requires high-voltage values larger than $30 kV$, so that $HV\_SetPoint \in 0 \cup [30  \: 90]  kV$. Consequently, the aforementioned limits are only specified at $30$ and $90 kV$ and over/under limits at other voltage levels can be calculated using simple linear interpolation between these values. All of these parameters are used for safety supervision purposes and are hence set in a conservative manner. In the next section, these parameters will be used to formulate and later formally verify the practical convergence property of the HVC controller to a new high-voltage setpoint.

\subsection{Properties for Formal Verification}\label{sec:properties}
In this section, the set of four properties that are to be formally verified is presented. Recognising that the HVC has a rather generic form of a feedback controller, it is notable that most of the properties in this section are rather natural and generic properties to be fulfilled by any feedback controller tracking a setpoint reference.

\paragraph{Property \textbf{P1}} 
To start with, it is natural to require that  the measured process value, which in the case of the HVC is dependent on time, \textit{t}, and denoted \textit{y(t) = HV\_Actual(t)}, should converge to the reference- or setpoint value, $r(t) = HV\_SetPoint(t)$. To formalize this, it is noted that both voltage signals are non-negative time-series and that convergence may be defined by setting
\begin{equation}
    e(t) = | r(t) - y(t) | = | HV\_SetPoint(t) - HV\_Actual(t)|, \label{eq:error_term}
\end{equation}
and equivalently considering (asymptotic) Lyapunov stability of the error term, $e(t)$, to origin. 

Taking the particular structure and dynamics of the HVC application as discussed previously into account, this setpoint convergence property can in practice  be decomposed into considering a 10 second time-interval directly after a new setpoint arrives, within which \textit{practical convergence} of \textit{HV\_Actual} to a narrow interval centered around the new external setpoint (\textit{HV\_SetPoint}) can be shown. To ease the notation and provide symmetry between the  Charge/Discharge modes, let
\begin{equation}\label{eq:settling_time_def}
    \tau = \tau_0 + \max(RampLimit, TauPeriod), 
\end{equation}
where $\tau_0$ denotes the time instance when a new setpoint arrives. Also conservatively, set the peak deviation from new \textit{HV\_SetPoint} as $30\%$ of the setpoint value. The width of this narrow interval, as well as schematic time changes and evolution of \textit{HV\_SetPoint} and \textit{HV\_Actual} are depicted in~\cref{fig:3_phases_of_PWM_HW}.  

This system-level property involves both hardware and software components and can be formally specified as follows:
\begin{enumerate}
    \item[\textbf{P1:}]\label[property]{prop:P1} Practical convergence of the actual system voltage, \textit{HV\_Actual}, to the external setpoint, \textit{HV\_SetPoint}:\\   
    \[\forall t \ge \tau \implies e(t) < 0.3 \times \max(HV\_SetPoint(t),1).\]
\end{enumerate}

\paragraph{Property \textbf{P2-P3}} 
To avoid residual effects and windup behaviours in the HVC, it is also reasonable to verify that both \textit{PWM\_Output} and the software internal representation of  \textit{HV\_SetPoint}, denoted \textit{mSetPoint}, are set to $0$ whenever the 24V power signal, and thereby the HVC-module, is switched off. Here, \textit{mSetPoint} is distinguished from \textit{HV\_SetPoint} which is a software extrinsic signal set \emph{a priori} by a human operator or application engineer. 

These two properties can be formulated as follows: 
\begin{enumerate}
    \item[\textbf{P2:}]\label[property]{prop:P2} That \textit{PWM\_Output} is set to $0$ whenever the 24V power signal is off:\\ \[24V\_Power = 0 \implies PWM\_Output = 0\]
    \item[\textbf{P3:}]\label[property]{prop:P3} That \textit{mSetPoint} is set to 0 when the 24V power signal is switched off:\\
    \[24V\_Power = 0 \implies mSetPoint = 0\]
\end{enumerate}

\paragraph{Property \textbf{P4}} 
Finally, in order to increase the confidence in the correctness of the model, it is customary to verify that the HVC state machine is not able to go into deadlock.
\begin{enumerate}
    \item[\textbf{P4:}]\label[property]{prop:P4} That the HVC software is not able to go into deadlock.
    
\end{enumerate}

These are the four properties that collectively need to be formally verified for the HVC application. System-level property \textbf{P1} is verified using our co-verification approach, which is the subject of~\cref{sec:co-verification}, while the verification of properties \textbf{P2-P4}, that only concern the software, is discussed in~\cref{sec:cover:software-properties}. Next we present an overview of the overall behaviour of the HVC software.

\subsection{Finite State Machine Overview}
\label{sec:finite}
\begin{figure}[t]
\centering
\includegraphics[width=.9\textwidth]{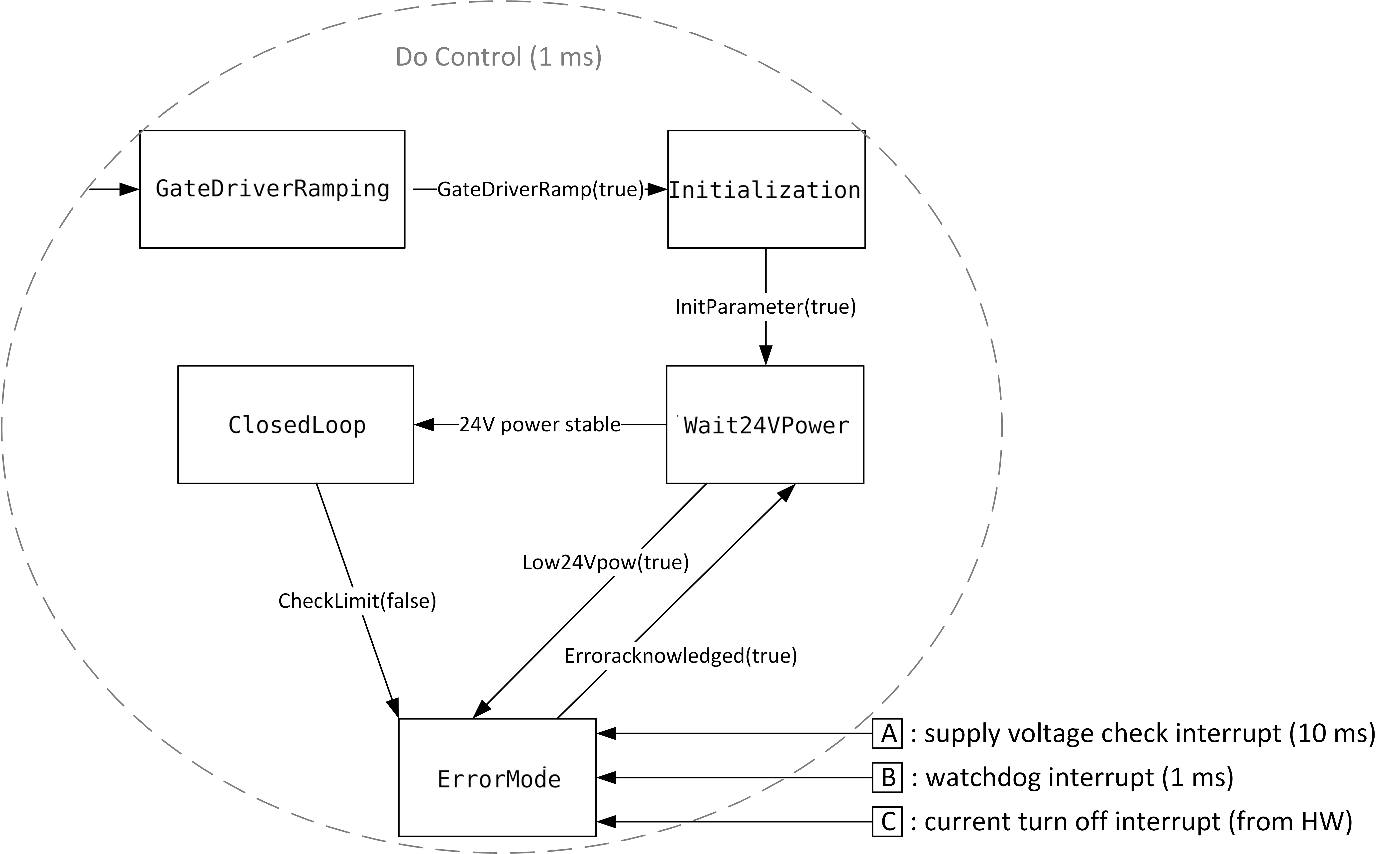}
\caption{\changed[C2]{Finite state diagram of the high-voltage Controller (HVC), showing its states and the conditions for transitioning between them. The \texttt{ClosedLoop} state is the ideal state for operation, and is where the controller is regulating the voltage in relation to the setpoint.}} \label{fig:HVC_state_machine}
\end{figure}

In order to perform model checking on the HVC, its functionalities were modelled as a finite state machine. This section presents the general finite state machine as depicted in~\cref{fig:HVC_state_machine}. This high-level state machine was given by ABB and then further detailed and modelled in RoboTool. This is the topic of~\cref{sec:cover:SW}. 

In the state \texttt{GateDriverRamping}, which is the state that the HVC first enters when it is switched on, the PWM duty-cycle is ramped up gradually to ensure stability and gradual increasing of current and voltage. Then, in the \texttt{Initialization} state, initial parameters are set, as well as upper and lower limits for the high-voltage.

After the \texttt{GateDriverRamping} and \texttt{Initialization} steps are successfully finished, the state machine enters the \texttt{Wait24VPower} state. When the HVC has 24V power switched on and is stable, the system enters the \texttt{ClosedLoop} state. This is the ideal state for operation, and is where the controller is regulating the voltage in relation to the setpoint. If the voltage breaches the upper or lower limits, the state machine moves from \texttt{ClosedLoop} to \texttt{ErrorMode}. 

There is also a possibility to enter \texttt{ErrorMode} from the \texttt{ClosedLoop} and \texttt{Wait24VPower} states, if certain variables are set or any watchdogs or interrupts are triggered. For instance, an interrupt is triggered if the supply voltage is below a certain threshold, and another is triggered if \textit{HV\_Actual} is above or below the upper and lower limits, respectively. Getting out of \texttt{ErrorMode} requires manual acknowledgement of the occurred errors. 

\section{Hardware/Software Co-Verification}
\label{sec:co-verification}
To reason about system properties, such as Property~\textbf{P1}, it is necessary to consider the behaviour of both software and hardware. We propose a novel approach, where properties are established by co-verification of models connected via platform mappings that relate the inputs and outputs of software and hardware, via sensors and actuators. Providing a crisp and systematic separation between hardware and software has some distinct advantages. To start with, with this approach, behavioural properties of individual models -- that may be established using domain-specific tools -- can be combined to support the verification of system properties. Also, our approach enables explicit recording and capturing of all dependencies and \changed[\C{2}]{relations} between hardware and software, which, when neglected, are implicitly assumed to be the identity mapping. Further, and as previously mentioned, the ongoing industrial trend is moving an increasing number of safety functions from physical hardware to software implementation. Still, the reliability, dependability and trust levels are very different between hardware and software components in an industrial robotic system. This framework hence sets the stage for more realistic and refined safety and risk handling procedures. Finally, the framework invites for combined approaches to modelling both discrete and continuous aspects in an integrated way while allowing the time-scale separation that typically exists between hardware and software components.

As an illustrative example, in our case study, the software is modelled in RoboChart, while the hardware is modelled in Simulink. RoboChart~\cite{RoboChartSoSym} is a domain-specific language for the model-based engineering of control software for robotics, that caters for timed and functional aspects. Its formal semantics is tailored for reasoning, namely using the CSP~\cite{Roscoe2010} model checker FDR~\cite{FDR}. However, it currently lacks facilities to specify the behaviour of the hardware. Simulink~\cite{Simulink}, on the other hand, is a \emph{de facto} standard for control engineering, typically used for dynamic simulation in the industrial setting of the HVC~\cite{thesis:morten,thesis:nina} system.

For modelling, we use Simulink and RoboTool~\cite{RoboChart,RoboStarTechnology2021}, that allows the graphical creation of RoboChart models, and for verification we use Simulink Design Verifier (SDV)~\cite{Simulink} and FDR. System Property~\textbf{P1} is co-verified by model-checking, using the formal semantics of the control software, as calculated by RoboTool, and an abstract specification of the hardware behaviour, as established using SDV. These are formalised in $tock$-CSP~\cite{Roscoe2010,Baxter2021}, the discrete-timed process algebraic semantics of RoboChart, for checking with FDR.

The complete system behaviour is considered at a suitable level of abstraction for verification by: (1) defining a platform mapping; (2) using a specification of the hardware that captures at an abstract level the relation between its inputs and outputs, as verified using SDV; (3) formalising these in $tock$-CSP. 
We depict the approach in~\cref{fig:HVC-framework} and explain it in detail in the next~\cref{sec:cover:framework}. In~\cref{sec:system-verification} we discuss the co-verification of system properties, modelling of the hardware and software, and the mechanisation in CSP of the overall framework. In~\cref{sec:cover:software-properties} verification of properties of the software is also discussed.

\subsection{Framework overview}
\label{sec:cover:framework}

\begin{figure}[tbp]
\centering%
\includegraphics[width=\textwidth]{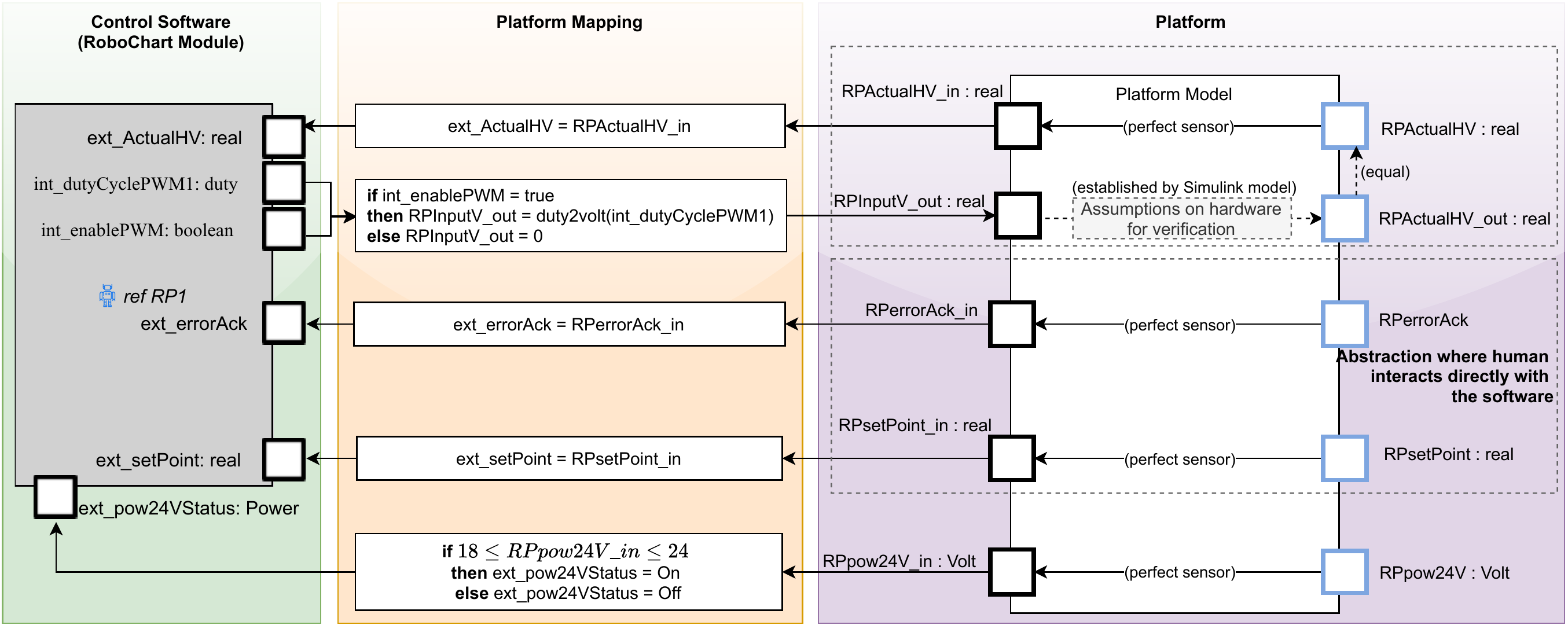}
\caption{Co-verification framework, with arrows indicating the direction of the information flow between inputs and outputs, of the software and hardware models. The platform mapping captures the relation between the software and hardware model on either side.} \label{fig:HVC-framework}
\end{figure}

In our framework, the software and hardware models are coupled via interfaces that capture their inputs and outputs, with connections between models specified via platform mappings. On the left-hand side of~\cref{fig:HVC-framework} we consider the interface of the HVC control software, defined as a robotic platform (\RC{RP1}) in RoboChart, that specifies the inputs and outputs as (possibly typed) events, indicated by solid boxes. 

On the right-hand side we have a high-level description of the hardware platform, that captures its sensors and actuators. In our abstraction of the HVC platform, that comprises the cascade in~\cref{HVC}, the hardware receives an input voltage, via \RC{RPInputV\_out}, and produces a high-voltage via \RC{RPActualHV\_out}.

We also annotate important assumptions about the hardware that are of relevance for analysis: sensors are perfect, and, in particular, the voltage produced via \RC{RPActualHV\_out} is assumed to be the same as that sensed via \RC{RPActualHV}. The relation between \RC{RPInputV\_out} and \RC{RPActualHV\_out} is established by the Simulink model as detailed in~\cref{sec:cover:SW}, but abstracted for verification, as explained later in~\cref{sec:cover:property}. The input signals \RC{RPerrorAck} and \RC{RPsetPoint} are an abstraction over the inputs available to a human operator, whose interaction with the software is realised via the platform.

The relation between the software and hardware model is specified by the platform mapping, as illustrated in the middle of~\cref{fig:HVC-framework}. It records how the inputs and outputs of the software are connected to sensors and actuators of the hardware platform, as realised by low-level code and physical interfaces. The mappings for the software inputs \RC{ext\_ActualHV}, \RC{ext\_errorAck} and \RC{ext\_newSetPoint} are trivial, as the software reads directly from these idealised sensors. The input \RC{ext\_pow24VStatus}, of type \RC{Power}, has the value \RC{On} if the reading from the hardware, via \RC{RPpow24V\_in} is between 18 and 24 Volts, and otherwise has the value \RC{Off}. 

The software outputs \RC{int\_dutyCyclePWM1} and \RC{int\_enablePWM} are used to determine whether a voltage is produced via \RC{RPInputV\_out}. If the value set via \RC{int\_enablePWM} is true, then the value of \RC{RPInputV\_out} is determined by the value of \RC{int\_dutyCyclePWM1}, otherwise it is 0. This captures the fact that the PWM needs to be enabled in order to produce a voltage. Here, the function \RC{duty2volt} maps a percentage, from 0 to 100\% to the range of the analog 0 to 10 voltage signal as previously mentioned in~\cref{sec:HVC}. 

More generally, when the connection between software and hardware is realised directly via reading and writing to registers of sensors and actuators, a platform mapping can be specified by a relation. As in the case of the input \RC{RPpow24V\_in}, a voltage is related to a discrete set of values \RC{On} and \RC{Off}. On the other hand, if inputs of the software are interpreted as event-driven interrupts, a more appealing approach is to define when an event is available~\cite{diagrammatic_RoboSim_models} using a predicate over the output of one or more sensors. Similarly, an input to an actuator may be constrained, for example, by a predicate over one or more software outputs.

\subsection{System verification}
\label{sec:system-verification}
Using the co-verification framework as illustrated in~\cref{fig:HVC-framework}, in this section we address the formal verification of system Property~\textbf{P1}. As described in~\cref{sec:properties}, it requires practical convergence of the high-voltage (\RC{RPActualHV\_out}) to the value of the setpoint as set by the user (\RC{RPsetPoint}). Since the software is modelled in RoboChart, and the hardware in Simulink, our pragmatic verification strategy consists of: (1) capturing \textbf{P1} as a specification in $tock$-CSP; (2) showing practical convergence of the hardware output \RC{RPActualHV\_out} in relation to its input \RC{RPInputV\_out} using SDV; (3) casting the result obtained from SDV as a $tock$-CSP specification; (4) checking with FDR that, when combined with the semantics of the RoboChart model, via a mechanisation of the framework depicted in~\cref{fig:HVC-framework}, \textbf{P1} is satisfied. That overall property~\textbf{P1} holds is justified by the timed process algebraic semantics of RoboChart and the abstract specification (2-3) as established using SDV, and captured in CSP. A full account of the CSP specifications for all of the properties considered in this paper can be found online\footnote{\url{https://github.com/UoY-RoboStar/hvc-case-study}}.

\paragraph{Formal Semantics} The formalism that we use, $tock$-CSP, is a dialect of the process algebra CSP, where the event $tock$ marks the passage of discrete time. As CSP adopts a reactive paradigm, interactions with the environment are specified using events, and that includes the passage of time in the case of $tock$-CSP. Importantly, it allows the specification of timed budgets and deadlines, and has a denotational semantics for refinement~\cite{Baxter2021}. Relevant for our work, the model checker FDR has tailored support for $tock$-CSP.

\begin{table}[t]\centering
  \begin{tabular}{|l|p{12.5cm}|}
    \hline
    \textbf{Process} & \textbf{Description}\\
    \hline
  	{\lstinline!SKIP!} & Terminating process\\
  	\hline
  	{\lstinline!WAIT(d)!} & Delay: terminates exactly after {\lstinline!d!} units of time have elapsed\\
  	\hline
    {\lstinline!STOP!} & Deadlock: no events are offered, but time can pass\\
    \hline
    {\lstinline!USTOP!} & Timelock: no events are offered and time cannot pass\\
    \hline
    {\lstinline!a -> P!} & Prefix operator: initially offers to engage in the event {\lstinline!a!} while permitting any amount of time to pass, and then behaves as {\lstinline!P!}\\
    \hline
    {\lstinline!if g then P else Q!} & Conditional: behaves as {\lstinline!P!} if the predicate {\lstinline!g!} is true, and otherwise as {\lstinline!Q!}\\
    \hline
   	{\lstinline!P [] Q!} & External choice of {\lstinline!P!} or {\lstinline!Q!} made by the environment\\
     \hline
   	{\lstinline!P ; Q!} & Sequence: behaves as {\lstinline!P!} until it terminates successfully, and, then it behaves as {\lstinline!Q!}\\
   	\hline
   	{\lstinline!P \ X!} & Hiding: behaves like {\lstinline!P!} but with all communications in the set {\lstinline!X!} hidden\\
   	\hline
   	{\lstinline!P |\ X!} & Project: behaves like {\lstinline!P!} but with all communications not in the set {\lstinline!X!} hidden\\
   	\hline
	{\lstinline!P ||| Q!} & Interleaving: {\lstinline!P!} and {\lstinline!Q!} run in parallel and do not interact with each other\\
	\hline
   	{\lstinline!P [| X |] Q!} & Generalised parallel: {\lstinline!P!} and {\lstinline!Q!} must synchronise on events that belong to the set {\lstinline!X!}\\
    \hline
     {\lstinline!P[[ c <- d]]!} & Renaming: replaces uses of channel {\lstinline!c!} with channel {\lstinline!d!} in {\lstinline!P!}\\
     \hline
     {\lstinline!P /\ Q!} & Interrupt: behaves as {\lstinline!P!} 
     until an event offered by {\lstinline!Q!} occurs, and then behaves as {\lstinline!Q!}\\
     \hline
     {\lstinline!P [| X |> Q!} & Exception: behaves as {\lstinline!P!} until {\lstinline!P!} performs an event in {\lstinline!X!}, and, then behaves as {\lstinline!Q!}\\
     \hline
    {\lstinline!TRUN(X)!} & Continuously offers the events in the set {\lstinline!X!} to the environment, while time can pass\\
    \hline
    {\lstinline!timed_priority(P)!} & Maximal progress: behaves as {\lstinline!P!} with internal behaviour given priority over $tock$, so that internal behaviour takes no time\\
    \hline
  \end{tabular}
  \caption{$tock$-CSP operators as used in a {\lstinline!Timed!} section, with basic processes at the top, followed by composite processes:~{\lstinline!P!} and {\lstinline!Q!} are metavariables that stand for processes, {\lstinline!a!}, {\lstinline!c!}, and {\lstinline!d!} for events, {\lstinline!g!} for a condition, and {\lstinline!X!} for a set of events.}
  \label{table:csp-operators}
\end{table}

In~\cref{table:csp-operators} we summarise the $tock$-CSP operators that we use in our work. To illustrate the notation we present a simple CSP specification of a one-place timed buffer in~\cref{fig:csp:example} written in \texttt{CSP$_\text{\texttt{M}}$}, the machine-readable version of CSP accepted by FDR. The declaration on~\cref{csp:example:data} introduces a named type \lstinline{data} whose values are defined by the set of integers between \lstinline{0} and \lstinline{2}. \Cref{csp:example:channel} declares two typed \lstinline{channel}s named \lstinline{in} and \lstinline{out}, that can be used as event constructors using the dot notation, for example, \lstinline{in.0} and \lstinline{out.1}. Related, the set of events generated by one or more event constructors can be specified as an enumerated set, so that, for example, \lstinline={|in|}= is equal to \lstinline={|in.0,in.1,in.2|}=, the set of all \lstinline{in} events.

Processes defined within a \lstinline{Timed} section\footnote{\url{https://cocotec.io/fdr/manual/cspm/definitions.html\#csp-timed-section}} (\crefrange{csp:example:Timed}{csp:example:Timed:end}) are interpreted to be timed: $tock$ events are automatically added to allow time to pass when waiting for interactions with the environment, and the passage of time is uniform across interacting processes. The function \lstinline{OneStep}, defined on \cref{csp:example:OneStep} to be \lstinline{0} for every event in its domain (indicated by the underscore in its signature), is passed as a parameter (\cref{csp:example:Timed}) to the \lstinline{Timed} section to ensure that no time is added implicitly after every synchronisation with the environment.

The behaviour of the timed buffer is that defined by the process \lstinline{Example} (\cref{csp:example:Example}) that behaves as \lstinline{TimedBuffer} with \lstinline{timed_priority} applied, a function over processes, provided by FDR to calculate the correct timed semantics~\cite{Baxter2021}. 
\lstinline{TimedBuffer} (\cref{csp:example:TimedBuffer}) initially offers to receive a \lstinline{data} value via a prefixing on the channel \lstinline{in}, and then offers an external choice (\lstinline{[]}) to the environment between accepting a new value, via the recursion on \lstinline{TimedBuffer}, or delaying the output of the current value, via prefixing on \lstinline{out} after the sequential composition (\lstinline{;}) with a delay of one time unit (\lstinline{WAIT(1)}). We observe that an external choice is not resolved by the passage of time, but rather by synchronisation on events.
Here, \lstinline{in?x} is syntactic sugar for accepting events \lstinline{in.x} where \lstinline{x} ranges over the type \lstinline{data}. The prefixing on \lstinline{out!x}
takes the value of \lstinline{x} as introduced into context by the prefixing on \lstinline{in?x}.

\begin{lstfloat}[t]
\begin{lstlisting}
nametype data = {0..2} (*@ \label{csp:example:data} @*)
channel in, out : data (*@ \label{csp:example:channel} @*)

OneStep(_) = 0 (*@ \label{csp:example:OneStep} @*)

Timed(OneStep) { (*@ \label{csp:example:Timed} @*)
  Example     = timed_priority(TimedBuffer) (*@ \label{csp:example:Example} @*)
  TimedBuffer = in?x -> (TimedBuffer [] (WAIT(1) ; out!x -> TimedBuffer)) (*@ \label{csp:example:TimedBuffer} @*)
} (*@ \label{csp:example:Timed:end} @*)
\end{lstlisting}

\vspace{-0.5ex}\caption{Example of a one-place buffer that is always prepared to receive a value, but which delays its output by one time unit.}\label{fig:csp:example}\vspace{-1.25ex}
\end{lstfloat}

\begin{lstfloat}[t]
\begin{lstlisting}
timed csp SpecP1 /@csp-begin@/
channel e, RPsetPoint : core_real (*@ \label{csp:SpecP1:channel} @*)

Timed(OneStep) {
 SpecP1 = timed_priority(Follow(s(3))) (*@ \label{csp:SpecP1:SpecP1} @*)
 Follow(d) = e?x -> (if x == 0 (*@ \label{csp:SpecP1:Follow} @*)
                     then Follow(d)
                     else ((ADeadline({|e|},{|e.0|},d) ; TRUN({|e.0|})) (*@ \label{csp:SpecP1:x-non-zero} @*)
                           /\ RPsetPoint?x -> Follow(d)) (*@ \label{csp:SpecP1:x-non-zero:end} @*)
                    )
             []
             RPsetPoint?x -> Follow(d) (*@ \label{csp:SpecP1:Follow:end} @*)
             
 -- Allows time to advance by 'd' units while events from 'S' are performed 
 -- until an event from 'S' that is in 'E' is performed.
 ADeadline(S,E,d) = EndBy(TRUN(S),d) [|E|> SKIP (*@ \label{csp:SpecP1:ADeadline} @*)
 
 -- Built into RoboTool: deadline for 'P' to terminate within 'd' units.
 EndBy(P,d) = P /\ (WAIT(d);USTOP) (*@ \label{csp:SpecP1:EndBy} @*)
}
/@csp-end@/
\end{lstlisting}
\vspace{-0.5ex}\caption{\label{csp:SpecP1}Specification for Property $\mathbf{P1}$ defined within a RoboChart assertions' process block named \lstinline{SpecP1}.}\vspace{-1.25ex}
\end{lstfloat}

\paragraph*{Specification}
Following the description of~\textbf{P1} in~\cref{sec:properties}, here we construct a discrete version in $tock$-CSP, as shown in the RoboChart \lstinline{timed csp} block named \lstinline{SpecP1} in~\cref{csp:SpecP1}. It declares two \lstinline{channel}s (\cref{csp:SpecP1:channel}), \lstinline{e} and \lstinline{RPsetPoint}, of type \lstinline{core_real}. The event \lstinline{e} is used to model the absolute difference between \RC{ActualHV\_out} and \RC{RPsetPoint}, so that the specification can capture the relation between changes in \RC{RPsetPoint} and the absolute difference.

The behaviour of \lstinline{SpecP1} is that of \lstinline{Follow}, a process with a single parameter \lstinline{d}, defined on \crefrange{csp:SpecP1:Follow}{csp:SpecP1:Follow:end} as an external choice (\lstinline{[]}) over accepting events \lstinline{e} or \lstinline{RPsetPoint}, via prefixing (\lstinline{?x ->}). Synchronisation on \lstinline{RPsetPoint}, with any value, or \lstinline{e}, with value 0, is followed by a recursion on \lstinline{Follow}. Whenever the event~\lstinline{e} carries a value that is not 0 (\crefrange{csp:SpecP1:x-non-zero}{csp:SpecP1:x-non-zero:end}), then~\lstinline{Follow} behaves as the process~\lstinline=ADeadline({|e|},{|e.0|},d)=, that ensures an event \lstinline{e} with a value of 0 can only be observed within~\lstinline{d} time units (instantiated as 3s for \lstinline{SpecP1}), and afterwards behaves as \lstinline=TRUN({|e.0|})=.
This behaviour can be interrupted (\lstinline=/\= on \cref{csp:SpecP1:x-non-zero:end}) at any time by a new~\lstinline{RPsetPoint}. We observe that for the purpose of model-checking the \RC{real}s, modelled by the type \lstinline{core_real}, are instantiated in the discrete domain 0 to 2, so here we consider the difference $e(t)$, encoded via the event~\lstinline{e}, to be 0, without loss of generality.

The auxiliary process \lstinline{ADeadline(S,E,d)}, defined on \cref{csp:SpecP1:ADeadline}, takes three parameters, two sets of events, \lstinline{S} and \lstinline{E}, where \lstinline{S} is expected to be a subset of \lstinline{E}, and a natural number \lstinline{d}. It continuously offers events in the set \lstinline{S}, but time can only advance \changed[\C{2}]{}by up to \lstinline{d} units, unless an event from the set \lstinline{E} happens, in which case the process terminates. It is defined using the exception operator of CSP (\lstinline{[|E|>}), where initially the behaviour is that of \lstinline{EndBy(TRUN(S),d)}, that continuously offers events in set \lstinline{S}, and allows time to advance by up to \lstinline{d} time units. Thus, within the exception, if \lstinline{TRUN(S)} performs an event that is in \lstinline{E}, then the process behaves as \lstinline{SKIP}, that terminates immediately. 

We observe that the auxiliary processes \lstinline{TRUN} and \lstinline{EndBy} are included with the RoboTool distribution for convenience. \lstinline{EndBy(P,d)}, reproduced on \cref{csp:SpecP1:EndBy}, is a deadline over process \lstinline{P} to terminate within \lstinline{d} time units. It behaves as \lstinline{P}, but because time is uniform in $tock$-CSP it requires the right-hand side process \lstinline{WAIT(d);USTOP} to agree on the passage of time. That process, however, is only prepared to let \lstinline{d} time units to pass before it timelocks, effectively imposing a deadline.
Next, we focus on the hardware model.

\subsubsection{Hardware Modelling and Verification in Simulink Design Verifier (SDV)}\label{sec:cover:HW}

Both the co-verification regime detailed in~\cref{sec:cover:framework}, as well as verification of the system-level properties, require a distinct and systematic separation between hardware and software components of the HVC system. \cref{fig:HVC-framework} provides an overview of this separation and the steps toward implementing this have been set forward in the ingress of~\cref{sec:co-verification}. To this end, the focus of this section is centered around hardware modeling, specification, abstraction and verification of hardware properties in SDV. All of these components are naturally combined in order to co-verify  system-level properties.

Simulink is widely adopted as a tool for traditional, input-driven simulation, and the modelling in SDV is similar to regular modelling used for simulation~\cite{Simulink}. 
SDV uses Prover Plug-In\textsuperscript{\tiny\textregistered} products from Prover\textsuperscript{\tiny\textregistered} Technology to do model-checking and prove the model properties~\cite{ProverAck}. It is built upon Gunnar Stålmarck's proof procedure, which uses tautology checks to prove that an assertion holds true in every possible interpretation~\cite{Stalmarck2000}. In Property Proving mode, SDV offers three different proof strategies, \texttt{Prove}, \texttt{FindViolation} and \texttt{ProveWithViolationDetection} where the latter is merely a serial combination of the two first mentioned. In this work, both \texttt{Prove} and \texttt{FindViolation} have been used. \texttt{Prove} performs an unbounded property proof, while \texttt{FindViolation} searches for property violations within the number of steps specified by the \texttt{Maximum violation steps} option, which specifies the maximum number of steps that SDV searches for property violations. Thus, verification with increasingly large \texttt{Maximum violation steps} will help to increase confidence in the property.

\paragraph{The Simulink Model}\label{sec:simulink_model}

\begin{figure}[!hbtp]
\includegraphics[width=\textwidth]{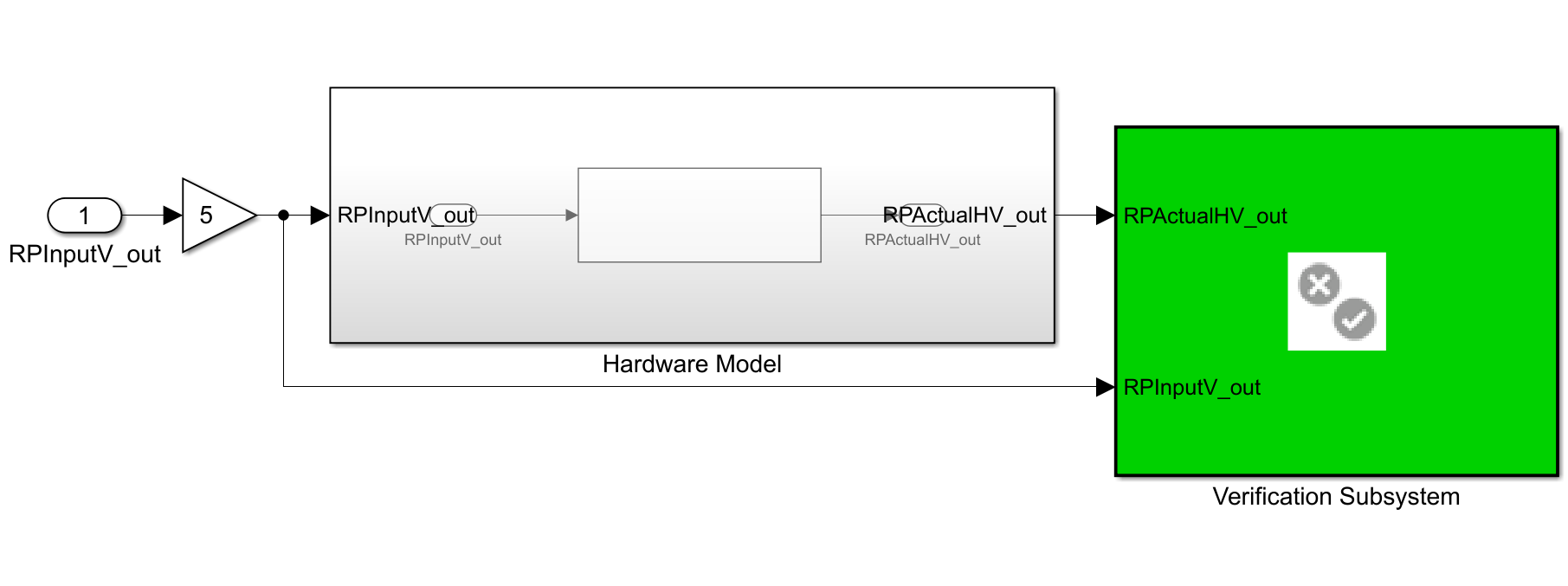}
\caption{Overview of the hardware verification model. The grey box includes the modelling of the cascade, while the green box contains the property for verification.} \label{fig:hpz}
\end{figure}

The hardware model in Simulink was created based on previous models found in~\cite{thesis:morten,thesis:nina}. These models have been validated both theoretically and empirically by several lab experiments, and correspond well to the real-world system. In order to do formal verification with SDV however, the model had to be converted from continuous to discrete time, since SDV does not support continuous time. In this process, in addition to converting transfer functions specified in continuous time using Laplace transform (S-domain) to discrete time Z-domain, some of the Simulink blocks specific to continuous time were replaced with their discrete counterparts. 
\cref{fig:hpz} shows the overview of the hardware verification model in SDV where the input/output signals, \emph{i.e.}, \RC{RPInputV\_out} and \RC{RPActualHV\_out} denote the same signals as previously introduced in~\cref{sec:cover:framework} and~\cref{fig:HVC-framework}. The mapping and transfer function between these two signals, and formal verification of certain hardware properties treated in this section, then correspond naturally to the extension and scope of the dashed grey box on the upper right side of~\cref{fig:HVC-framework}. 

It is noteworthy that, the input, \RC{RPInputV\_out}, ranges over the discrete domain, $\{0, 1, 2\}$, and is multiplied by a constant factor 5 in Simulink, effectively corresponding to having the set of possible values of $\{0, 5, 10\}$ volt being fed into the PWM hardware model. This means that \RC{duty2volt} maps a percentage, from 0 to 100\% to the entire range of the analogue $[0 \; 10]$ voltage signal as previously mentioned in~\cref{sec:HVC}. It also implies that the convergence results obtained in this section using $\{0, 5, 10\}$ volt as input, will also be valid for the real PWM hardware system. This is because its input set,  $[0 \; 10]$, is a superset of $\{0, 5, 10\}$.

\begin{figure}[!htbp]
\centering
\includegraphics[width=\textwidth]{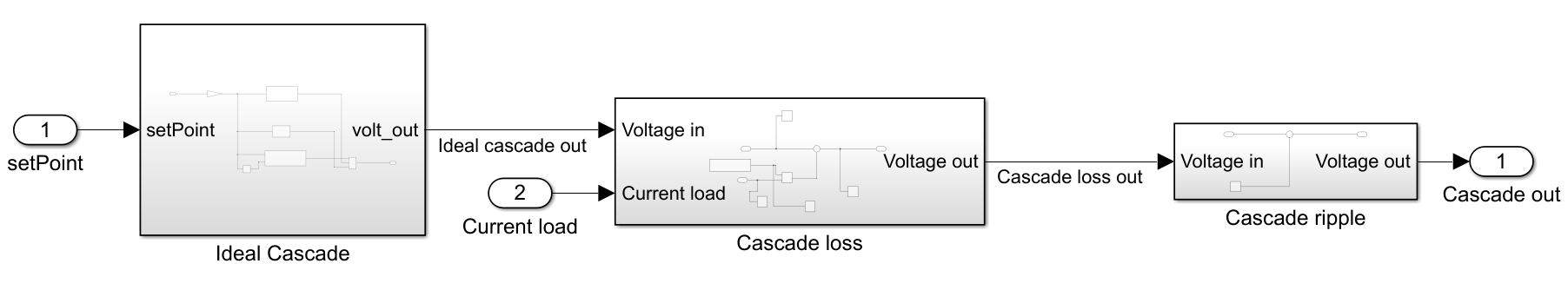}
\caption{Simulink model of the total CW-cascade hardware complementing the ideal model with loss and ripple terms~\cite{thesis:nina}.} \label{fig:CW_cascade_simulink_model}
\end{figure}

The test data used to create the model was collected from structured experiments running at many different HV setpoints, frequencies, distances and number of stages in the CW-cascade, providing a rich data-set to represent how the actual hardware will behave in the real environment. As detailed in~\cite{thesis:morten,thesis:nina} and depicted in~\cref{fig:CW_cascade_simulink_model}, the Simulink model will, in addition to the ideal transfer function, have two additional terms describing the cascade loss and ripple effects. Using the Matlab System Identification Toolbox, state-space models and transfer-functions are fitted to the lab test data to provide the best description of the PWM hardware dynamics; both during the charge\changed[\C{2}]{} and discharge modes of operation. The resulting transfer functions and model components in continuous time can be seen in~\cref{fig:CW_cascade_simulink_model}. Additionally, a Simulink model describing the bell-cup inside the applicator that will effect the electrical field at a plane at a given distance, $d$, from the paint robot, has been derived in~\cite{thesis:morten} and used here.

\begin{figure}[!htbp]
\centering
\includegraphics[width=\textwidth]{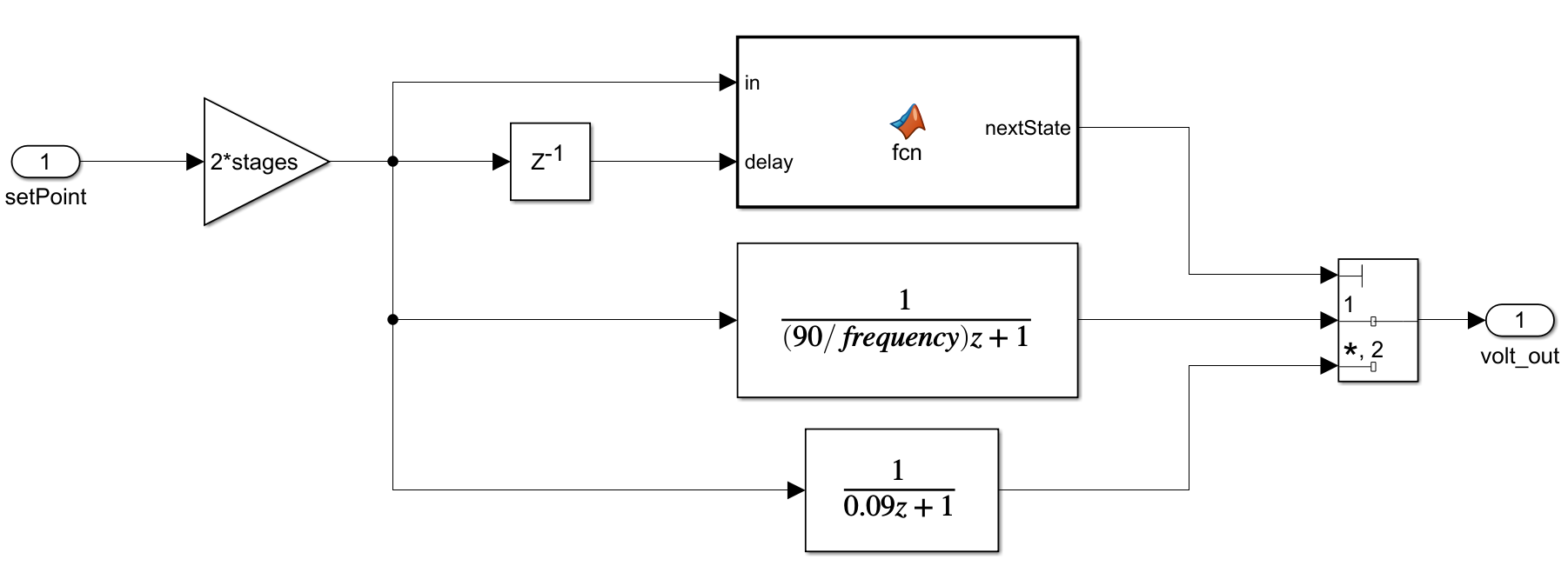}
\caption{Simulink model of the ideal cascade including the mode model selector and the two transfer functions describing the charging\changed[\C{2}]{} and discharging modes respectively~\cite{thesis:nina}.} \label{fig:CW_transfer_fcns}
\end{figure}

In order to be able to formally verify the system-level property, \textbf{P1}, the mapping and relational properties between \RC{RPInputV\_out} and \RC{RPActualHV\_out}, effectively describing the hardware, is needed. This allows us to obtain a well-defined ``closed circuit'' \changed[\C{2}]{}mapping between all of the components in the co-verification framework of~\cref{fig:HVC-framework}. To this end, the System Identification Toolbox was used to model the transfer function describing the relation between these two signals. This resulted in a standard, second order transfer function model:
\begin{equation}
    G(s) = \frac{K_p}{(1+T_{p1}s)(1+T_{p2}s)},
\end{equation}
having the following specific parameters
\begin{align*}
    K_p &= 1.1196 \\                  
    T_{p1} &= 0.087821  \\               
    T_{p2} &= 0.02042.      
\end{align*}
This model was then analysed in Simulink with particular attention to time dynamics, stability and convergence properties as defined by, \emph{e.g.}, rise and settling-time. Of particular interest in the following, is the settling time, $t_s$, which was found to be $t_s = 0.3668 s$.

\paragraph{Formal Verification of Hardware Properties}
Based on the developed Simulink model, next, we will be verifying a low-level property that will then be used in the co-verification scheme in order to verify Property~\textbf{P1}.

Referring back to the definition of Property~\textbf{P1}, the error term~\cref{eq:error_term} as well as the notion of \emph{practical convergence} in~\cref{sec:properties}, the following hardware property will be considered and verified in this section:

$\mathbf{P_{HW}}:$ Practical convergence of actual hardware output voltage, \RC{RPActualHV\_out}, to the hardware input signal, \RC{RPInputV\_out}, within settling time, $t_s$:   
\begin{align}
 &\forall t \ge t_{sp} + t_s \implies  \nonumber \\    
 &E(t) = |\RC{RPInputV\_out}(t) - \RC{RPActualHV\_out}(t) | - 0.15 \times \max(\RC{RPInputV\_out}(t),1) \le 0. \label{eq:HW_convergence_property} 
\end{align}

Here, $t_{sp}$ denotes the time instance where an update to the  input, \RC{RPInputV\_out}, is received in PWM hardware. It is notable that while Property~\textbf{P1} is a system-level property, involving both software and hardware components, Property~$\mathbf{P_{HW}}$ as defined by~\cref{eq:HW_convergence_property} serves as a low-level hardware property. Another important distinction, stems from the difference in value between the settling time, $t_s = 0.3668s$, in~\cref{eq:HW_convergence_property} and $\tau = 3s$ in~\cref{eq:error_term} which implies that $\mathbf{P_{HW}}$ incrementally contributes towards fulfilment of corresponding equations to verify the overall convergence Property~\textbf{P1}. This fact also underlies and explains the difference in peak deviation limit (0.15 and 0.3 respectively) between the two properties.

\begin{figure}[!htbp]
\centering
\includegraphics[width=\textwidth]{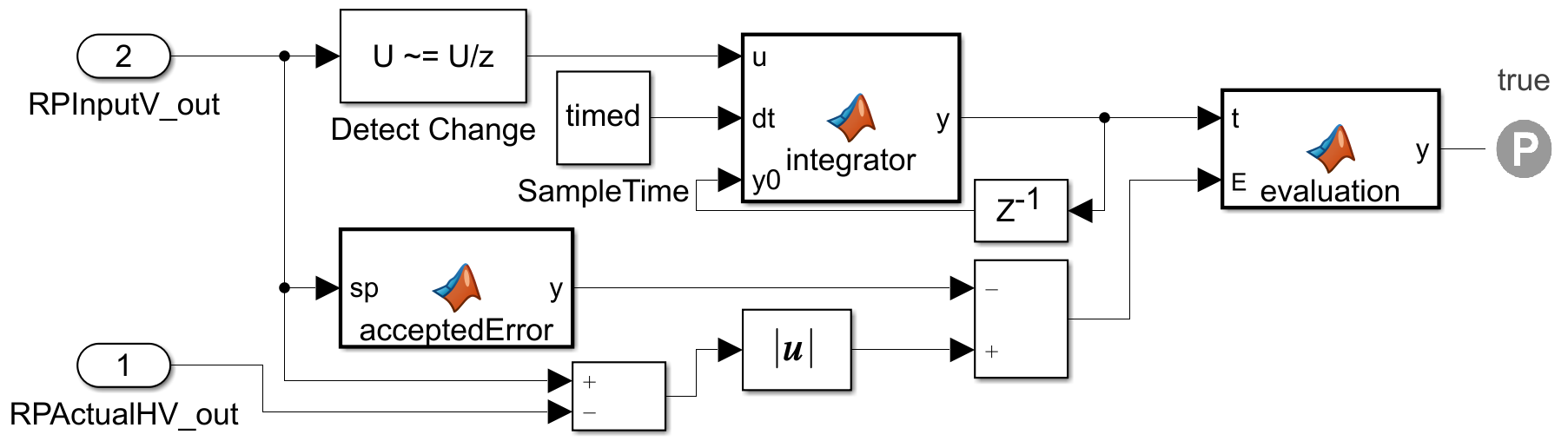}
\caption{SDV implementation of the PWM hardware convergence property, $\mathbf{P_{HW}}$ as detailed in~\cref{eq:HW_convergence_property}.} \label{fig:simulinkspecification}
\end{figure}

The Simulink implementation to verify this property lies within the green Verification Subsystem in~\cref{fig:hpz} and has been depicted in~\cref{fig:simulinkspecification}.
The upper part containing the \RC{Detect Change} block and an integrator function, works as a timer that is reset every time there is a change in \RC{RPInputV\_out}. This in order to capture the $t \ge t_{sp} + t_s$ constraint in~\cref{eq:HW_convergence_property}. The lower part takes the absolute value of the error between \RC{RPInputV\_out} and \RC{RPActualHV\_out} and subtracts the accepted error, which is set to $0.15 \times \max(\RC{RPInputV\_out}(t),1)$. Finally, the last function on the right, denoted evaluation, gives out false if~\cref{eq:HW_convergence_property} is not fulfilled at any time instance, $t \ge t_{sp} + t_s$. Otherwise it gives out true. This is verified with the proof assumption block, which shows if the property is fulfilled or violated.

After creating the model and the specification, the \texttt{Prove} strategy was used in order to verify the property. It was run both using MATLAB online and on a Windows laptop with Intel\textsuperscript{\textregistered} Core\textsuperscript{\textcopyright} i5 CPU @ 2.71GHz. The online version and the desktop version were both used in order to run two verifications in parallel, to see if one would produce a result faster than the other. However, after running both continuously for 10 days without producing a result, the verification was manually terminated, with the assumption that the complexity had resulted in a state-space explosion that made Simulink unable to \changed[\C{2}]{verify} the specification. It was instead decided to gain increased confidence in the verification by using \texttt{FindViolation} with increasing \texttt{Maximum violation steps}.
The results of the verification by using \texttt{FindViolation} can be seen in~\cref{tab:hwRes}. The \texttt{Maximum violation steps} option was gradually increased, until reaching the maximum value of 2,147,483,647, which is the maximum value for data type \texttt{int32}. As seen in the table, SDV was able to prove that the property was valid within bound in all cases.

\begin{table}[!htbp]
\centering
\begin{tabular}{llll}
\hline
\multicolumn{1}{|c|}{\textbf{Maximum violation steps}} & \multicolumn{1}{|c|}{\begin{tabular}[|c|]{@{}c@{}}\textbf{Fixed-step size}\\\textbf{(fundamental sample time)}\end{tabular}} & \multicolumn{1}{|c|}{\textbf{Result}} & \multicolumn{1}{|c|}{\textbf{Elapsed time}}\\
\hline
\multicolumn{1}{|l|}{1,000} & \multicolumn{1}{l|}{$1e^{-6}$}    & \multicolumn{1}{l|}{Valid within bound}  & \multicolumn{1}{l|}{0:47:49}  \\
\hline
\multicolumn{1}{|l|}{1,000,000} & \multicolumn{1}{l|}{$1e^{-6}$}    & \multicolumn{1}{l|}{Valid within bound}  & \multicolumn{1}{l|}{0:46:44}  \\ 
\hline
\multicolumn{1}{|l|}{1,000,000,000} & \multicolumn{1}{l|}{$1e^{-6}$}    & \multicolumn{1}{l|}{Valid within bound}  & \multicolumn{1}{l|}{0:47:15}  \\
\hline
\multicolumn{1}{|l|}{2,147,483,647} & \multicolumn{1}{l|}{$1e^{-6}$}    & \multicolumn{1}{l|}{Valid within bound}  & \multicolumn{1}{l|}{0:47:15}  \\ 
\hline
\end{tabular}
\cprotect\caption{Results of the verification of the hardware, using \texttt{FindViolation} and different values for \texttt{Maximum violation steps}.}
\label{tab:hwRes}
\end{table}

\subsubsection{Software Modelling in RoboChart}\label{sec:cover:SW}

\begin{figure}[tp]
\centering
\includegraphics[width=.65\textwidth]{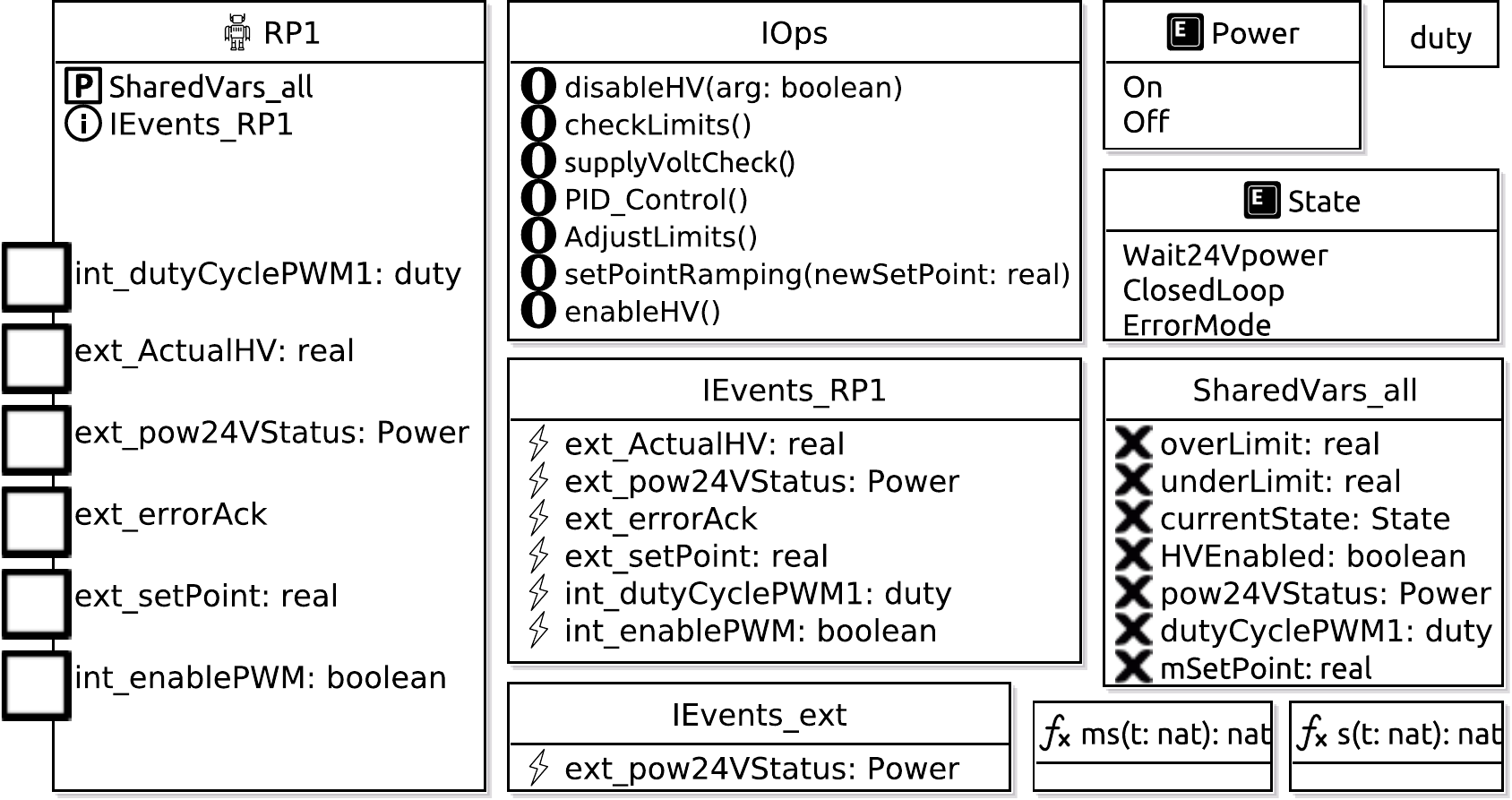}
\cprotect\caption{RoboChart model components: robotic platform (\protect\includegraphics[height=8pt]{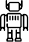} \RC{RP1}), interfaces (named \RC{IOps}, \RC{IEvents\_RP1}, \RC{IEvents\_ext}, and \RC{SharedVars\_all}), enumerated (\RC{Power} and \RC{State}) and given (\RC{duty}) data types. \protect\includegraphics[height=8pt]{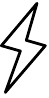} is an event, \protect\includegraphics[height=8pt]{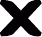} is a variable, and \includegraphics[height=8pt]{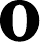} is associated with an operation. \includegraphics[height=8pt]{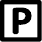} is used to record that an interface is provided, while \includegraphics[height=8pt]{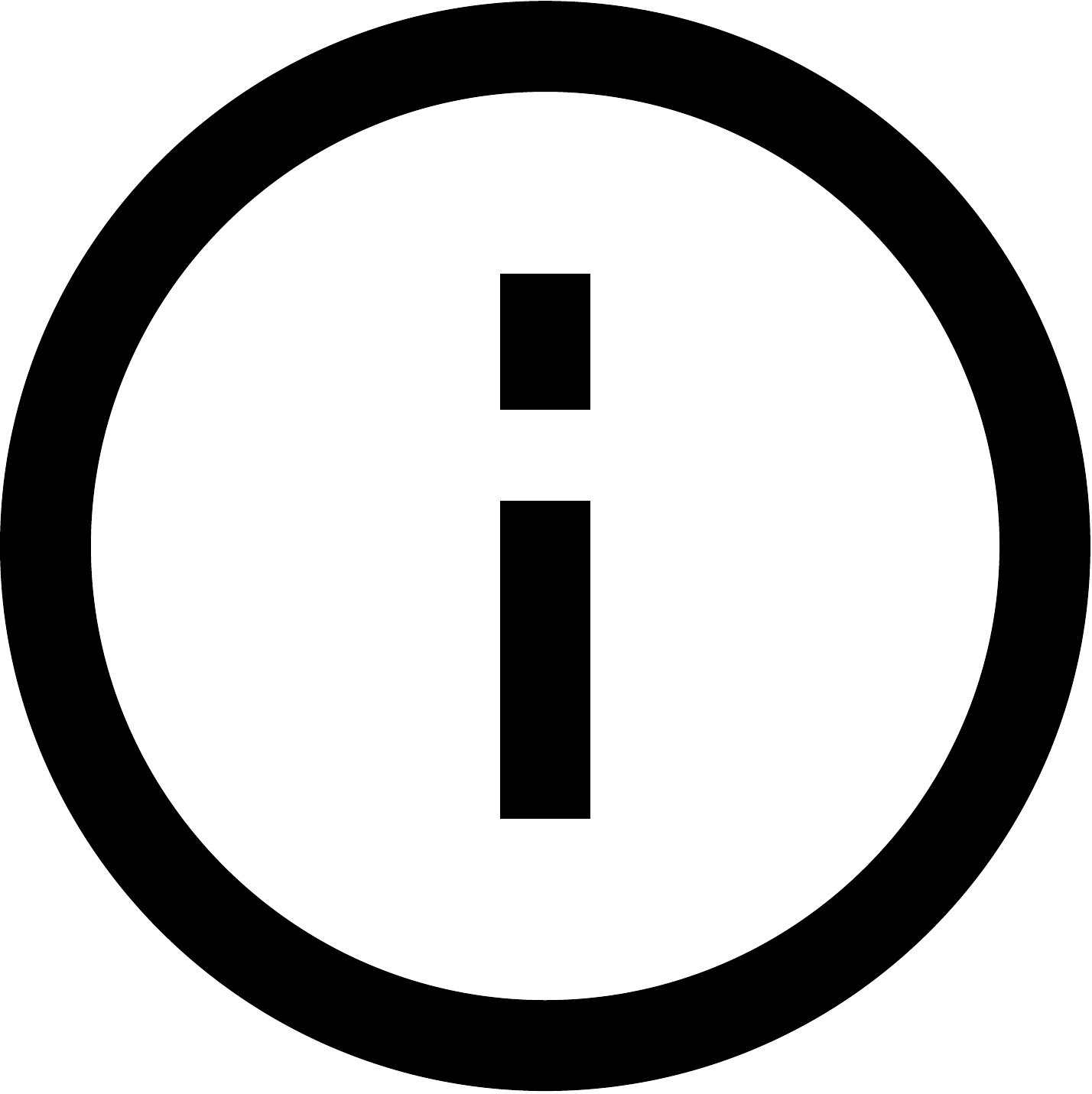} is a used interface.}
\label{fig:RoboTool}
\end{figure}

In this section, we present the RoboChart model of the software, that is a formalisation of the sketch previously shown in~\cref{fig:HVC_state_machine}. The robotic platform (RP1) -- a specification of the services available to the software in terms of variables, events and operations -- is fully specified in~\cref{fig:RoboTool}. Its events are defined in the interface \RC{IEvents\_RP1}. \RC{RP1} also provides the interface \RC{SharedVars\_all}, that declares all of the shared variables that are used in the model. The interface \RC{IOps} specifies the signature of the software operations that are used, and defined, in the RoboChart model. In addition to employing built-in data types, such as \RC{real}s, \RC{nat}urals, and \RC{boolean}s, three data types are declared: the enumerated types \RC{Power} and \RC{State}, and the given type \RC{duty}. Two functions \RC{ms} and \RC{s} are used to construct time units corresponding to milliseconds and seconds, respectively. RoboChart adopts the type system of Z~\cite{WD96,ISOZ}. For a full account of the language and its formal semantics we refer the reader to~\cite{RoboChart,RoboTool,RoboStarTechnology2021}. Here, we describe the RoboChart constructs as we use them to model our example.

\begin{figure}[!h]
\centering
\includegraphics[width=.8\textwidth]{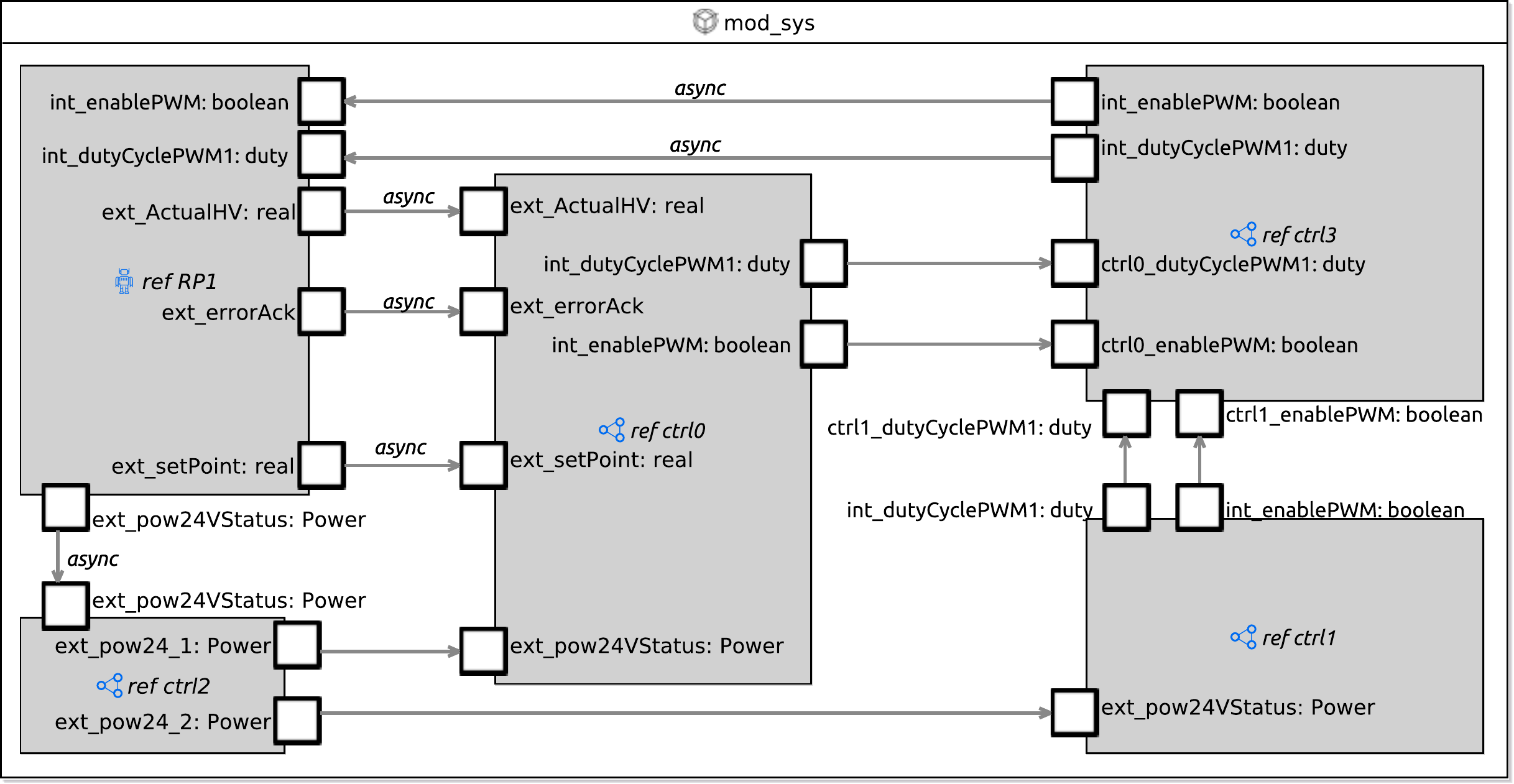}
\caption{RoboChart module \RC{mod\_sys} defining the connections between controllers and the robotic platform. Controller \RC{ctrl0} contains the main \RC{State\_machine}, a recast in RoboChart of the state machine presented in~\cref{fig:HVC_state_machine}. The watchdogs have been combined into one state machine, defined inside controller \RC{ctrl1}. Controller \RC{ctrl2} relays the event \RC{ext\_pow24VStatus} to controllers \RC{ctrl0} and \RC{ctrl1}, while controller \RC{ctrl3} is used for relaying the events \RC{int\_enablePWM} and \RC{int\_dutyCyclePWM1} to \RC{RP1}.}\label{fig:Module}
\end{figure}

\begin{figure}[tpb]
    \centering
    \includegraphics[width=0.65\textwidth]{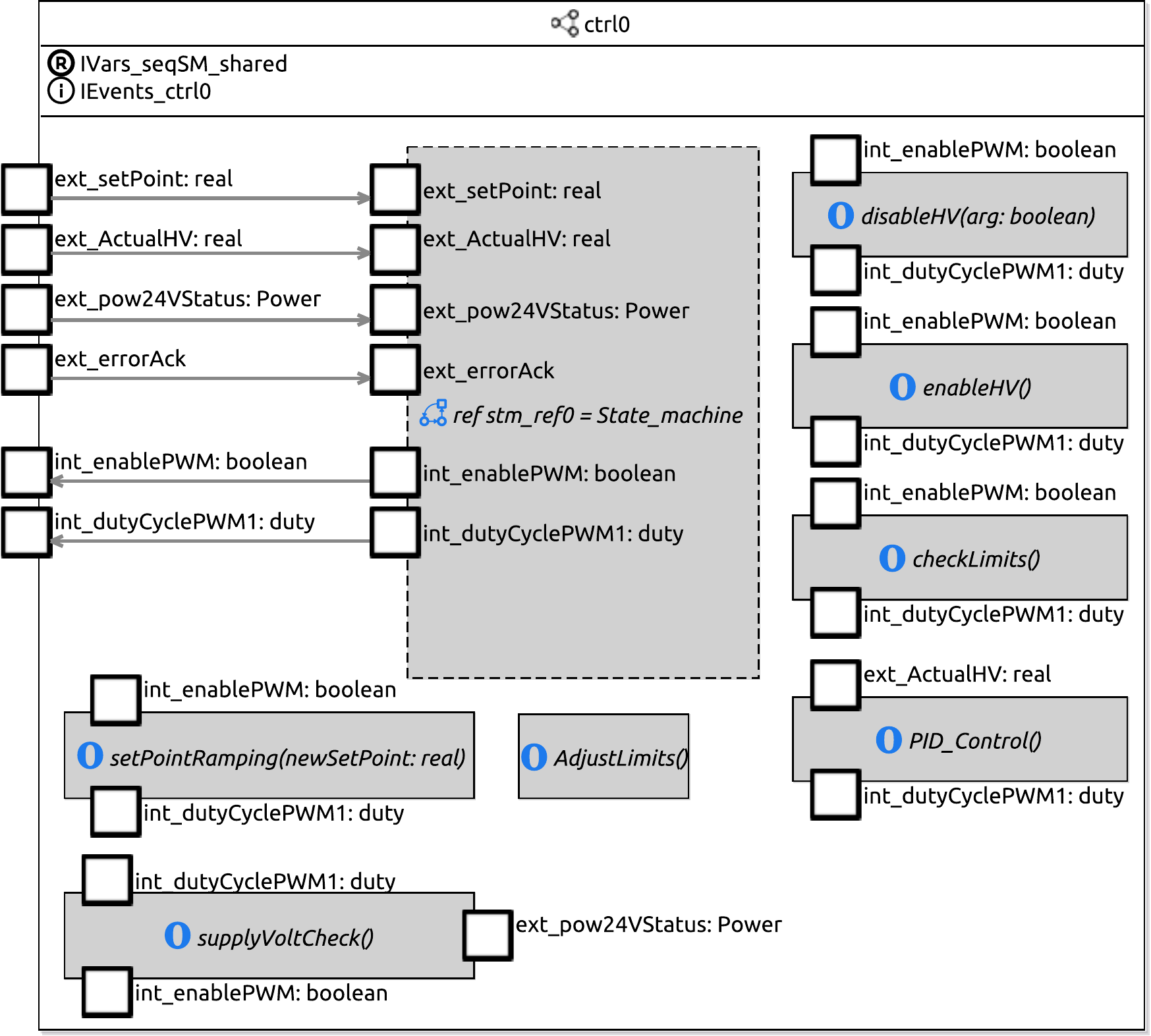}
    \caption{Controller \RC{ctrl0}, showing its inputs and output connections to \RC{Sate\_machine}, and references of operations.}
    \label{fig:ctrl0}
\end{figure}

\paragraph{Module and Controllers} The top-level component of the software model is defined by the RoboChart module \RC{mod\_sys}, shown in~\cref{fig:Module}. It associates the robotic platform with four controllers (\RC{ctrl0-3}), that capture specific behaviours. Controller \RC{ctrl0}, shown in~\cref{fig:ctrl0}, contains the main \RC{State\_machine}, that is a recast of that presented in~\cref{fig:HVC_state_machine}, \RC{ctrl1} captures the behaviour of the watchdogs, and controllers \RC{ctrl2-3} are used to relay events. Controller \RC{ctrl2} relays the input event \RC{ext\_pow24VStatus} from \RC{RP1} to controllers \RC{ctrl0-1}, and \RC{ctrl3} relays the output events \RC{int\_dutyCyclePWM1} and \RC{int\_enablePWM} from \RC{ctrl0} and \RC{ctrl1} to \RC{RP1}, as RoboChart event connections are point-to-point. Due to their simple nature, the full definition of all controllers is deferred to an on-line report\footnote{\url{https://robostar.cs.york.ac.uk/case_studies/hvc/}}. In RoboChart, connections with the platform are always asynchronous, indicated by the keyword \RC{\emph{async}}, as interactions with the platform cannot be refused, only ignored~\cite[p.3110]{RoboChartSoSym}. The connections between all controllers \RC{ctrl0-3}, however, are set as synchronous.

\begin{figure}[tp]
\begin{center}
\includegraphics[width=\textwidth]{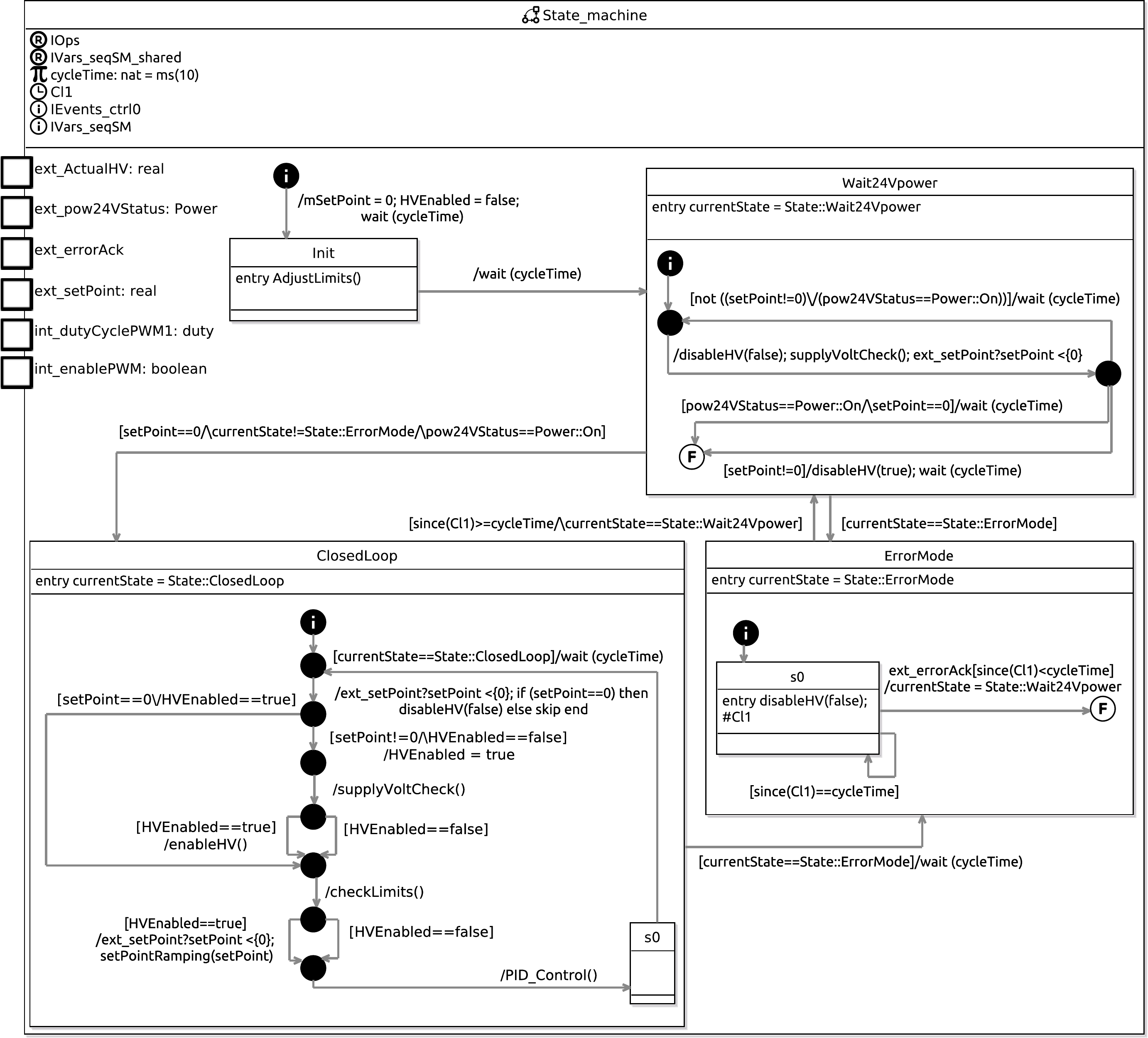}
\caption{Main \RC{State\_machine} corresponding to that of~\cref{fig:HVC_state_machine} recast in RoboChart.}
\label{fig:robotoolfull}
\end{center}
\end{figure}

\paragraph{State Machine}
The core behaviour of the HVC controller is captured by the \RC{State\_machine} in~\cref{fig:robotoolfull}. In RoboChart, state machines are self-contained  by explicitly stating the required ({\includegraphics[height=9pt]{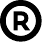}}) variables and operations, and the used ({\includegraphics[height=9pt]{Figures/Icons/used.pdf}}) events. In this case, \RC{State\_machine} requires the software operations declared in \RC{IOps}, and the shared variables in \RC{IVars\_seqSM\_shared}. It also declares: local variables via the interface \RC{IVars\_seqSM}, a constant \RC{cycleTime} with a default value of 10 milliseconds, and a clock ({\includegraphics[height=9pt]{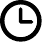}}) \RC{Cl1}. It uses the events of interface \RC{IEvents\_ctrl0}, that are also explicitly listed on the left-hand side of~\cref{fig:robotoolfull}.

\begin{figure}[tbp]
\begin{center}
\includegraphics[width=.65\textwidth]{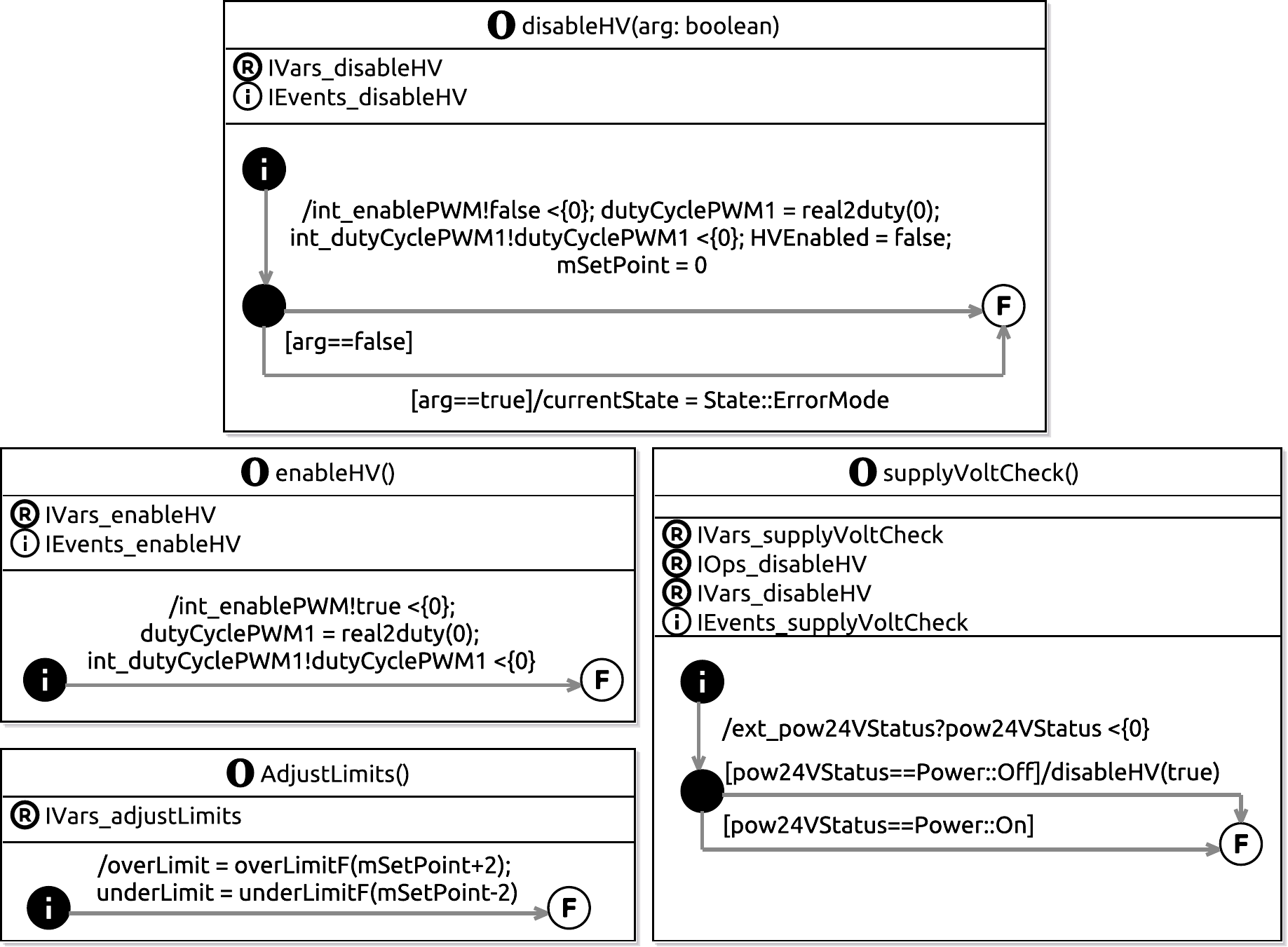}
\vspace{-1ex}\caption{Software operations modelled in RoboChart with their behaviour defined by state machines: \RC{disableHV} disables the high-voltage by writing \RC{false} to \RC{int\_enablePWM} and setting the duty cycle to zero via \RC{int\_dutyCyclePWM1}; \RC{enableHV} enables the high-voltage; \RC{AdjustLimits} calculates voltage limits, \RC{overLimit} and \RC{underLimit}, based on the current value of \RC{mSetPoint}; and \RC{supplyVoltCheck} checks whether the 24V power is stable via the input \RC{ext\_pow24VStatus}, and if unstable calls \RC{disableHV}.}
\label{fig:operations}
\end{center}\vspace{-1ex}
\end{figure}

The execution flow of \RC{State\_machine} starts at the initial junction, followed by a transition whose action, specified after the dash (\RC{/}), initializes the value of the variables \RC{mSetPoint} and \RC{HVEnabled}, by assigning \RC{0} and \RC{false} in sequence (\RC{;}), respectively. It then \RC{wait}s for \RC{cycleTime} units before entering state \RC{Init}. This initial delay is a simplification of the \texttt{GateDriverRamping} behaviour depicted in~\cref{fig:HVC_state_machine}, which does not concern the properties of interest for verification. In state \RC{Init} there is an entry action that calls the software operation \RC{AdjustLimits} which calculates the value of variables \RC{overLimit} and \RC{underLimit} and is defined by a state machine as shown in~\cref{fig:operations}. The required variables of \RC{AdjustLimits}, as listed in interface \RC{IVars\_adjustLimits}, are provided by \RC{State\_machine} in the context of the call to \RC{AdjustLimits}, effectively sharing the state.

After the initialization is complete, the execution proceeds to the composite state \RC{Wait24Vpower} on the next cycle. Its entry action explicitly records that the state has been entered by setting the variable \RC{currentState}. The transition to \RC{ClosedLoop} is only enabled when the current value of~\RC{setPoint} is \RC{0}, the 24V power is stable (\RC{pow24VStatus==Power::On}), and the \RC{ErrorMode} is not activated, as indicated by the transition's guard. The body of \RC{Wait24Vpower} monitors the relevant inputs periodically as part of the cycle of transitions between the junctions. Firstly, the operations~\RC{disableHV} and \RC{supplyVoltCheck}, as defined in~\cref{fig:operations}, are called. \RC{disableHV} disables the high-voltage, while \RC{supplyVoltCheck} checks the input~\RC{ext\_pow24VStatus} and updates the value of the variable \RC{pow24VStatus}. Secondly, the value of the variable \RC{setPoint} is also updated via a reading (\RC{ext\_setPoint?setPoint}) through event \RC{ext\_setPoint}, with a deadline~(\RC{<\{0\}}) of zero time units. In RoboChart budgets and deadlines must be specified explicitly, and so here the deadline indicates that the reading takes a negligible amount of time.

The critical phase of HVC operation is captured in state \RC{ClosedLoop}, that controls the PWM. Initially the user-defined setpoint, \RC{ext\_SetPoint},  is read into the variable~\RC{setPoint}. If the value is zero, then \RC{disableHV} is called to ensure that the high-voltage is disabled. Afterwards, if the value is non-zero and the high-voltage has not been enabled yet (\RC{HVEnabled==false}), \RC{HVEnabled} is set to \RC{true}, the supply voltage is checked by calling \RC{supplyVoltCheck()}, and the high-voltage is enabled by calling \RC{enableHV}. While the high-voltage is enabled, the internal setpoint (recorded in variable \RC{mSetPoint}) is adjusted by calling \RC{setPointRamping(setPoint)}. The PWM duty-cycle is adjusted by \RC{PID\_Control} that outputs a percentage via \RC{int\_dutyCyclePWM1}, according to the difference between \RC{mSetPoint} and \RC{ActualHV}, the measured high-voltage via the input \RC{ext\_ActualHV}. In state \RC{s0} of \RC{ClosedLoop}, the flow of execution may be interrupted by transitioning to \RC{ErrorMode} when \RC{currentState} is set to \RC{State::ErrorMode}. The error can be acknowledged via the event \RC{ext\_errorAck} within the current \RC{cycleTime}, after which there is a transition to \RC{Wait24Vpower}.

The variable \RC{currentState} may be set to \RC{State::ErrorMode} by calling \RC{disableHV(true)}, either while in \RC{Wait24Vpower}, or from within operations \RC{checkLimits} or \RC{supplyVoltCheck}, that checks whether the input 24V power is stable. The latter is called regularly in states \RC{ClosedLoop} and \RC{Wait24Vpower} of \RC{State\_machine}, and also by the watchdog, which, as will be  explained next, is modelled in another state machine.

\begin{figure}[tbp]
\begin{center}
\includegraphics[width=.5\textwidth]{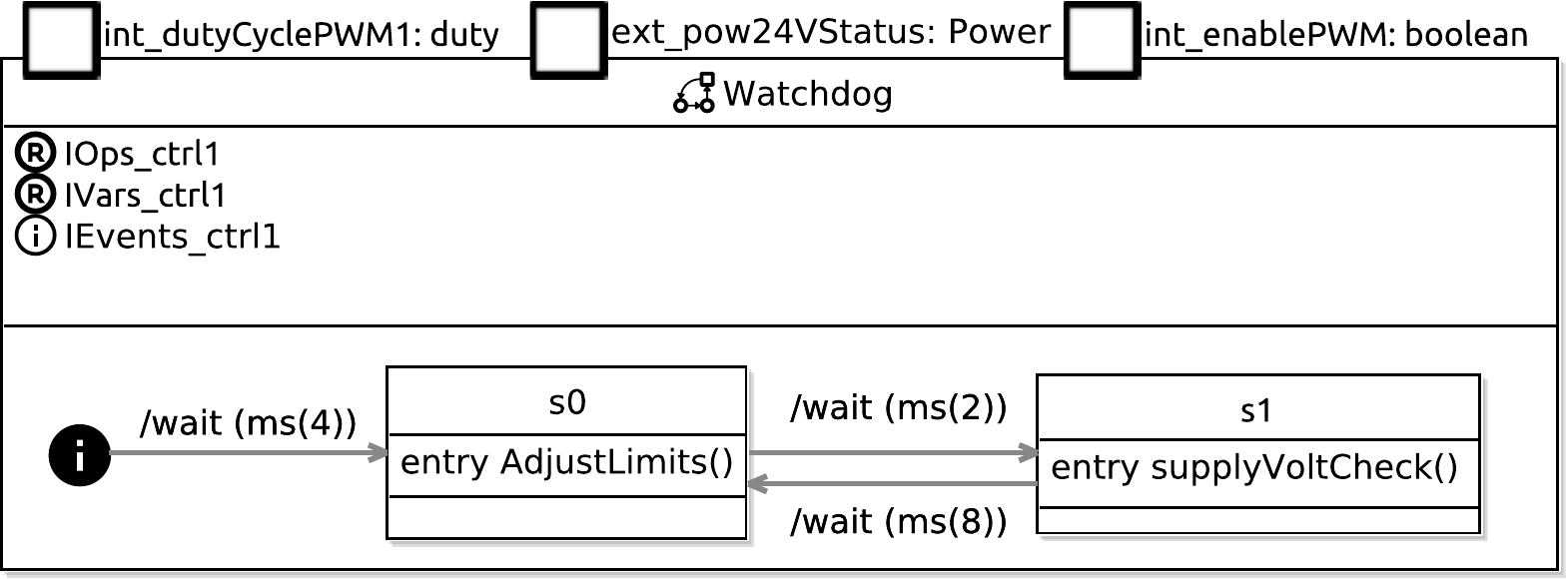}
\vspace{-1ex}\caption{\RC{Watchdog} state machine. Initially there is a delay of 4 milliseconds before engaging in a cyclic behaviour. First \RC{AdjustLimits} is called, and after 2 milliseconds \RC{supplyVoltCheck} is called, with same behaviour repeated after a delay of a further 8 milliseconds. The call to \RC{supplyVoltCheck} can disable the high-voltage if the 24V power is not stable, as shown in~\cref{fig:operations}.}
\label{fig:watchdog}
\end{center}
\end{figure}

\paragraph{Watchdog}
The watchdog, shown in~\cref{fig:watchdog}, executes, over time, in alternation with the main \RC{State\_machine}, that executes on a 10 millisecond cycle, as specified by the constant \RC{cycleTime}. Therefore, the watchdog's behaviour is initially delayed by 4 milliseconds.
In state \RC{s0} there is a call to \RC{AdjustLimits()}, and 2 milliseconds later, the operation \RC{supplyVoltCheck()} is called. We observe that the transition between \RC{s1} and \RC{s0} takes 8 milliseconds, and it is during this time that \RC{State\_machine} actually executes its cyclic behaviour.

\begin{lstfloat}[thb]
\begin{lstlisting}
timed csp HVC_Platform /@csp-begin@/
nametype Power_Voltage = {0,24}  -- Data type used to characterise input RPpow24V

channel RPActualHV_out  : core_real 		          -- 'Actuator' output (*@\label{csp:HVC-Platform:RPActualHV_out}@*)
channel RPInputV_out    : core_real 		          -- Output from software via mapping 
channel RPActualHV, RPActualHV_in  : core_real 	-- Sensor input and platform mapping
channel RPpow24V, RPpow24V_in  : Power_Voltage 	-- Sensor input and platform mapping
channel RPsetPoint_in  : core_real 	            -- Sensor platform mapping
channel RPerrorAck, RPerrorAck_in               -- Untyped input and platform mapping (*@\label{csp:HVC-Platform:RPerrorAck}@*)

channel get_HV, set_HV, change : core_real      -- Used for abstraction of the hardware

Timed(OneStep) {
 HVC_Platform = timed_priority((HV(0)[[ get_HV <- RPActualHV_out ]] (*@\label{csp:HVC-Platform:HVC_Platform}@*)
                                [| {|set_HV|} |] 
                                HVC_Hardware)\{|set_HV|} (*@\label{csp:HVC-Platform:HVC_Platform:end}@*)
                              )
    
 HV(x) = set_HV?nv -> HV(nv) [] get_HV!x -> HV(x)(*@\label{csp:HVC-Platform:HV}@*)
    
 HVC_Hardware = (Detector [| {| change, get_HV |} |] StatefulEvolution)\{|change, get_HV|}(*@\label{csp:HVC-Platform:HVC_Hardware}@*)
    
 Detector = RPInputV_out?nv -> get_HV?x -> (if nv != x
                                            then (change!nv -> Detector) 
                                            else Detector)
    
 StatefulEvolution = (Evolution [| {| change |} |] HV(0)[[ set_HV <- change ]])(*@\label{csp:HVC-Platform:StatefulEvolution}@*)
    
 Evolution = change?x -> ((WAIT(ms(370)); set_HV!x -> Evolution) [] Evolution)
}
/@csp-end@/
\end{lstlisting}%
\vspace{-0.5ex}\caption{\label{csp:hvc-platform} Mechanisation of the \lstinline{HVC_Platform}.}\vspace{-1.25ex}
\end{lstfloat}

\subsubsection{Framework Mechanisation}
\label{sec:cover:mechanisation}

Having developed models of the software and hardware, in this section we mechanise the co-verification framework outlined in~\cref{fig:HVC-framework} with the aim of verifying system Property~\textbf{P1}. We start by defining a CSP process that captures Property~$\mathbf{P_{HW}}$. This is followed by the complete mechanisation of the platform and its mapping, and the composition with the semantics of the RoboChart model, as calculated by RoboTool.

\paragraph{Platform}
The hardware platform is specified within the \lstinline{csp} block named~\lstinline{HVC_Platform}, shown in~\cref{csp:hvc-platform}. It defines, first of all, the \texttt{CSP$_\text{\texttt{M}}$} events of the sensors and actuators (\crefrange{csp:HVC-Platform:RPActualHV_out}{csp:HVC-Platform:RPerrorAck}), following the naming conventions of~\cref{fig:HVC-framework}.
The process \lstinline{HVC_Platform} is a discrete, and reactive, model of the hardware, constructed from the Property~$\mathbf{P_{HW}}$ established in~\cref{sec:cover:HW}. It is defined as a parallel composition (\lstinline{[| |]} on \crefrange{csp:HVC-Platform:HVC_Platform}{csp:HVC-Platform:HVC_Platform:end}), synchronising on event~\lstinline{set_HV}, of~\lstinline{HV(0)}, that models the current value of the high-voltage, and~\lstinline{HVC_Hardware}, that captures how the value of~\lstinline{RPActualHV_out} may change over time in response to changes in~\RC{RPInputV\_out}. The process~\lstinline{HV(x)}, defined on \cref{csp:HVC-Platform:HV}, offers the event~\lstinline{set_HV} to change the value, and the event~\lstinline{get_HV} to query the current value~\lstinline{x}. It is specialised on \cref{csp:HVC-Platform:HVC_Platform} as \lstinline{HV(x)[[ get_HV <- RPActualHV_out ]]} by renaming the event~\lstinline{get_HV} to~\lstinline{RPActualHV_out}. The event~\lstinline{set_HV} is hidden ({\lstinline!\!}), as it is an artefact of the CSP model.

The evolution of the value available via~\RC{RPActualHV\_out} is modelled by the process~\lstinline{HVC_Hardware}. It is defined on \cref{csp:HVC-Platform:HVC_Hardware} as the parallel composition of the process~\lstinline{Detector}, synchronising on the events \lstinline{change} and \lstinline{get_HV}, and the process~\lstinline{StatefulEvolution}. The latter models how changes to the voltage evolve over time, while~\lstinline{Detector}, named analogously to the SDV block in~\cref{fig:simulinkspecification}, models how an input via~\RC{RPInputV\_out} may affect the behaviour. First it offers to receive a new value \lstinline{nv} via \lstinline{RPInputV_out}, and then synchronises with \lstinline{StatefulEvolution} on \lstinline{get_HV} to query the current value \lstinline{x} being targeted. If the value is different, it synchronises on \lstinline{change} with value \lstinline{nv}, otherwise it behaves as \lstinline{Detector}.

The core of the hardware property $\mathbf{P_{HW}}$ is abstractly captured by the process \lstinline{StatefulEvolution}. It is defined on \cref{csp:HVC-Platform:StatefulEvolution} as the parallel composition of \lstinline{Evolution}, synchronising on event \lstinline{change}, with \lstinline{HV(0)} where the channel \lstinline{set_HV} is renamed to \lstinline{change}. \lstinline{Evolution} accepts a \lstinline{change} event at anytime, and afterwards waits 370 milliseconds, a conservative natural approximation of the settling time of the Simulink model of the hardware, before synchronising on \lstinline{set_HV}, which is used to update the high-voltage, whose value is available via \RC{RPActualHV\_out}, as modelled by the process \lstinline{HV(0)} in \lstinline{HVC_Platform}. Thus, a change via \lstinline{RPInpuV_out} leads to a change in the value available via \lstinline{RPActualHV_out} over time, mirroring property $\mathbf{P_{HW}}$ as established in~\cref{sec:cover:HW}. 
Next, we describe the mechanisation of the platform mapping.

\begin{lstfloat}[th]
\begin{lstlisting}
timed csp PlatformMapping /@csp-begin@/
-- Function that maps the duty cycle percentage to the value expected by RPInputV_out. 
duty2volt(x) = if member(x,{0..19}) then 0(*@\label{csp:platform-mapping:duty2volt}@*)
                  else (if member(x,{20..60}) then 1 
                        else (if member(x,{61..100}) then 2 else 0))(*@\label{csp:platform-mapping:duty2volt:end}@*)
Timed(OneStep) {
 PlatformMapping = timed_priority((RPInputV_out!0 -> PWM_Map(false)) ||| Pow24_Map(true))(*@\label{csp:platform-mapping:PlatformMapping}@*)

 -- Platform mapping of duty cycle and PWM to RPInputV_out.
 PWM_Map(pwm) = (*@\label{csp:platform-mapping:PWM_Map}@*)
  mod_sys::int_enablePWM.out?x -> (*@\label{csp:platform-mapping:PWM_Map:x-false}@*)
   ((if x == false then RPInputV_out!0 -> SKIP else SKIP) ; PWM_Map(x))(*@\label{csp:platform-mapping:PWM_Map:x-false:end}@*)
  []
  mod_sys::int_dutyCyclePWM1.out?x -> (*@\label{csp:platform-mapping:PWM_Map:pwm-true}@*)
   ((if pwm == true then RPInputV_out!duty2volt(x) -> SKIP else SKIP); PWM_Map(pwm))(*@\label{csp:platform-mapping:PWM_Map:end}@*)

 -- Platform mapping from RPpow24V_in to ext_pow24VStatus.
 Pow24_Map(pwr) = (*@\label{csp:platform-mapping:Pow24_Map}@*)
  RPpow24V_in?x:{x | x <- Power_Voltage, (x <= 24 and x >= 18)} -> Pow24_Map(true)(*@\label{csp:platform-mapping:RPpow24V_in:true}@*)
  []
  RPpow24V_in?x:{x | x <- Power_Voltage, (not (x <= 24 and x >= 18))} -> Pow24_Map(false)(*@\label{csp:platform-mapping:RPpow24V_in:false}@*)
  []
  (if pwr==true
   then mod_sys::ext_pow24VStatus.in!Power_On -> Pow24_Map(pwr)
   else mod_sys::ext_pow24VStatus.in!Power_Off -> Pow24_Map(pwr))(*@\label{csp:platform-mapping:Pow24_Map:end}@*)
}
/@csp-end@/
\end{lstlisting}%
\vspace{-0.5ex}\caption{\label{csp:platform-mapping} Mechanisation of the Platform Mapping.}\vspace{-1.5ex}
\end{lstfloat}

\paragraph{Platform Mapping}
The process~\lstinline{PlatformMapping}, defined on \cref{csp:platform-mapping:PlatformMapping} of the \lstinline{csp} block in~\cref{csp:platform-mapping}, captures the non-trivial mapping between \RC{int\_dutyCyclePWM1}, \RC{int\_enablePWM}, and \RC{RPInputV\_out}, and between \RC{RPpow24V\_in} and \RC{ext\_pow24VStatus}.
It is an interleaving (\lstinline{|||}) of two processes, \lstinline{Pow24V_Map}, that models the mapping between \RC{RPpow24V\_in} and the software input \RC{ext\_extPow24VStatus}, and the prefixing on \lstinline{RPInputV_out} with value 0 followed by the behaviour of \lstinline{PWM_Map}. Here the prefixing initializes the hardware with value zero. \lstinline{PWM_Map}, defined on \crefrange{csp:platform-mapping:PWM_Map}{csp:platform-mapping:PWM_Map:end}, models the mapping between the outputs \RC{int\_dutyCyclePWM1} and \RC{int\_enablePWM}, and the input to the platform \RC{RPInputV\_out}.

\lstinline{PWM_Map} is parametrised to keep track of whether the PWM has been turned on or off. The first process in the external choice (\crefrange{csp:platform-mapping:PWM_Map:x-false}{csp:platform-mapping:PWM_Map:x-false:end}) allows this value to be toggled depending on whether \lstinline{int_enablePWM.out} is received with value \lstinline{false}, in which case the value zero is passed to the platform via \lstinline{RPInputV_out}, and otherwise there is a recursion on \lstinline{PWM_Map(x)} with the updated value of \lstinline{x}. In the second process (\crefrange{csp:platform-mapping:PWM_Map:pwm-true}{csp:platform-mapping:PWM_Map:end}), values received via \lstinline{int_dutyCyclePWM1.out} are passed to the platform via \lstinline{RPInputV_out}, mapped via the function \lstinline{duty2volt}, if the value of \lstinline{pwm} is currently \lstinline{true}. This function, defined on \crefrange{csp:platform-mapping:duty2volt}{csp:platform-mapping:duty2volt:end}, maps a percentage to a voltage, which, as previously discussed in~\cref{sec:cover:framework}, encodes three possible values.

The CSP process \lstinline{Pow24_Map} is defined analogously on \crefrange{csp:platform-mapping:Pow24_Map}{csp:platform-mapping:Pow24_Map:end} to model the mapping between the sensor of the 24V voltage, and the input \RC{ext\_pow24VStatus} of the software, whereby a value between 18 and 24 is considered as \RC{On} and otherwise as \RC{Off}. The value that \lstinline{x} can take on the prefixings on the channel \lstinline{RPpow24V_in}, on lines \ref{csp:platform-mapping:RPpow24V_in:true} and \ref{csp:platform-mapping:RPpow24V_in:false}, is constrained using set comprehensions. This concludes the non-trivial mappings, which are used in the definition of the overall system next.

\begin{lstfloat}[th]
\begin{lstlisting}
timed csp MappedSystem /@csp-begin@/
Timed(OneStep) {
-- Defined over the semantics of the RoboChart module 'mod_sys' calculated by RoboTool.
Software = mod_sys::O__(0,ms(10),1) [[ mod_sys::ext_ActualHV.in <- RPActualHV_in,(*@\label{csp:mapped-system:Software}@*)
                                       mod_sys::ext_errorAck.in <- RPerrorAck_in,
                                       mod_sys::ext_setPoint.in <- RPsetPoint_in ]]
Software_PMap = (*@\label{csp:mapped-system:Software_PMap}@*)
 (Software
  [|{|mod_sys::int_enablePWM.out,
      mod_sys::int_dutyCyclePWM1.out,
      mod_sys::ext_pow24VStatus|}|]
  PlatformMapping)
  \{|mod_sys::int_enablePWM.out,mod_sys::int_dutyCyclePWM1.out,mod_sys::ext_pow24VStatus|}(*@\label{csp:mapped-system:Software_PMap:end}@*)

MappedSoftware = Software_PMap[[ RPActualHV_in <- RPActualHV,(*@\label{csp:mapped-system:MappedSoftware}@*)
                                 RPerrorAck_in <- RPerrorAck,
                                 RPpow24V_in   <- RPpow24V,
                                 RPsetPoint_in <- RPsetPoint ]](*@\label{csp:mapped-system:MappedSoftware:end}@*)

MappedSystem = timed_priority(MappedSoftware(*@\label{csp:mapped-system:MappedSystem}@*)
                              [| {| RPInputV_out, RPActualHV |} |]
                              (HVC_Platform[[ RPActualHV_out <- RPActualHV_out, 
                                              RPActualHV_out <- RPActualHV]])
                              \{|RPInputV_out,RPActualHV|})(*@\label{csp:mapped-system:MappedSystem:end}@*)
}
/@csp-end@/
\end{lstlisting}%
\vspace{-0.5ex}\caption{\label{csp:mapped-system}Complete mechanisation of the Platform Mapping.}\vspace{-1.25ex}
\end{lstfloat}

\paragraph{Mapped System}
The complete system, as envisioned in~\cref{fig:HVC-framework} is modelled by the process \lstinline{MappedSystem} within the \lstinline{csp} block of~\cref{csp:mapped-system}.
It is defined on \crefrange{csp:mapped-system:MappedSystem}{csp:mapped-system:MappedSystem:end} as the parallel composition of \lstinline{MappedSoftware} and \lstinline{HVC_Platform}, defined in~\cref{csp:hvc-platform}, synchronising on the events \lstinline{RPInputV_out} and \lstinline{RPActualHV}. Here \lstinline{HVC_Platform} is relationally renamed~\cite[p. 105]{Roscoe2010}, so that the event~\lstinline{RPActualHV_out} is both an output of the platform and also a sensor input, with the same value, via~\lstinline{RPActualHV}, as depicted in~\cref{fig:HVC-framework}. The hiding on~\lstinline{RPInputV_out} and \lstinline{RPActualHV} completes the abstraction.

The process \lstinline{MappedSoftware}, defined on \crefrange{csp:mapped-system:MappedSoftware}{csp:mapped-system:MappedSoftware:end}, captures the connections between the composition of the platform mapping and the software, as established by~\lstinline{Software_PMap}, and the platform, by renaming the events of the former to the latter. The sensors of the platform, in particular, are assumed to be perfect, and so in this abstraction the functional renaming is a record of their ideal functional behaviour.

\lstinline{Sofware_PMap}, defined on \crefrange{csp:mapped-system:Software_PMap}{csp:mapped-system:Software_PMap:end}, captures the composition of the RoboChart CSP semantics, and the~\lstinline{PlatformMapping}, as defined earlier in~\cref{csp:platform-mapping}. It is a parallel composition of the processes \lstinline{Software} and \lstinline{PlatformMapping}, synchronising on \lstinline{int_enablePWM.out}, \lstinline{int_dutyCyclePWM1}, and \lstinline{ext_pow24VStatus}, corresponding to events of the RoboChart model. The process \lstinline{Software}, which explicitly instantiates the RoboChart model semantics, is defined analogously to~\lstinline{MappedSoftware}, whereby the trivial mappings are captured via renaming. The hiding on \cref{csp:mapped-system:Software_PMap:end} completes the abstraction. 

The process~\lstinline{mod_sys::O__(0,ms(10),1)}, used on \cref{csp:mapped-system:Software}, is an explicit instantiation of the \texttt{CSP$_\text{\texttt{M}}$} semantics of \RC{mod\_sys}, automatically calculated by RoboTool, where \lstinline{0} is a default identifier, \lstinline{ms(10)} is the value of constant \RC{cycleTime} of \RC{State\_machine}, and \lstinline{1} the value of constant \RC{rampStep} of operation \RC{setPointRamping}. Events in the CSP semantics of RoboChart are named according to the model hierarchy, where \lstinline{::}~is a delimiter, and have a parameter \lstinline{in} or \lstinline{out} to indicate whether an event is an input or output.

\begin{lstfloat}[th]
\begin{lstlisting}
timed csp SpecP1 /@csp-begin@/
Timed(OneStep) {
 SpecP1 = timed_priority(Follow(s(3)))
 Follow(d) = e?x -> (if x == 0 
                     then Follow(d)
                     else ((ADeadline({|e|},{|e.0|},d) ; TRUN({|e.0|})) 
                           /\ RPsetPoint?x -> Follow(d))
                    )
             []
             RPsetPoint?x -> Follow(d)
} 
/@csp-end@/

// Actual check for P1
timed assertion P1 : ImplP1 refines SpecP1 /@in the traces model@/(*@\label{csp:SpecP1-verification:P1}@*)
\end{lstlisting}%
\vspace{-0.5ex}\caption{\label{csp:SpecP1-verification}Recall of \lstinline{SpecP1} as previously defined in~\cref{csp:SpecP1} and the RoboChart assertion for verification.}\vspace{-1.25ex}
\end{lstfloat}

\subsubsection{Formal Verification of System-level Property P1}
\label{sec:cover:property}
With the framework outlined in~\cref{fig:HVC-framework} mechanised in CSP, in this section we address the verification of Property~\textbf{P1}. Its specification in CSP, described in~\cref{sec:cover:framework}, is reproduced in~\cref{csp:SpecP1-verification} for convenience.
Verification of Property~\textbf{P1} is stated as a refinement \lstinline{assertion P1} \lstinline{/@in the traces model@/} of CSP (\cref{csp:SpecP1-verification:P1}), that ensures safety~\cite[p.36]{Roscoe2010}. That is, an implementation \lstinline{P} \lstinline{refines} \lstinline{Spec}, if, and only if, every behaviour of the implementation is a behaviour permitted by the specification. While RoboChart \lstinline{/@assertions@/} are translated by RoboTool into \CSPM~refinement assertions, they can also directly reference elements of the RoboChart model to facilitate the checking of basic properties, such as deadlock freedom and termination~\cite[p. 3129]{RoboChartSoSym}. For \lstinline{assertion P1}, \lstinline{SpecP1} is the specification and \lstinline{ImplP1} is the implementation. As previously discussed in~\cref{sec:cover:framework}, \lstinline{SpecP1} is stated in terms of a new event \lstinline{e}, that is not part of~\cref{fig:HVC-framework} but useful to specify Property \textbf{P1} in terms of the absolute difference between the value of output \RC{RPActualHV\_out} and input \RC{RPsetPoint}. To facilitate verification, process \lstinline{ImplP1}, is defined next to relate events \lstinline{e} and \lstinline{RPActualHV_out}, and \lstinline{RPsetPoint}, based on the process \lstinline{MappedSystem}, previously defined in~\cref{csp:mapped-system}.

\paragraph{System Interface for Verification of P1}
We observe that in~\lstinline{SpecP1}, the event \lstinline{RPsetPoint} is used as an interrupt, which emerges naturally in the reactive CSP setting. However, the event~\lstinline{RPsetPoint} as used so far in the definition of~\lstinline{MappedSystem} models readings of a sensor, that can be performed periodically despite no change in the actual value. Therefore, to relate \lstinline{SpecP1} and \lstinline{ImplP1}, in~\cref{csp:ImplP1} we define a suitable mapping for the \lstinline{RPsetPoint} event. We also capture the relationship between the event \lstinline{e} and the current value of both the setpoint and the actual high-voltage, as required for the comparison with \lstinline{SpecP1}.

\begin{lstfloat}[!ht]
\begin{lstlisting}
timed csp ImplP1 /@csp-begin@/
channel int_RPsetPoint : core_real
abs_diff(x,y) = if (x-y >= 0) then (x - y) else (y - x) 

Timed(OneStep) {
 -- Assumptions required of the system input and outputs for P1 to hold.
 Assumption_RPerrorAck = STOP(*@\label{csp:ImplP1:Assumption_RPerrorAck}@*)
 Assumption_RPpow24V = RPpow24V!24 -> STOP(*@\label{csp:ImplP1:Assumption_RPpow24V}@*)
 Assumption_SetPoint = EndBy(RPsetPoint.0 -> SKIP,0); WAIT(ms(22)); RPChange(*@\label{csp:ImplP1:Assumption_SetPoint}@*)
 RPChange = RPsetPoint?x -> WAIT(s(1)); RPChange(*@\label{csp:ImplP1:RPChange}@*)
 
 -- Composition of MappedSystem and model of Assumptions 1-2.   
 SystemP1 = (*@\label{csp:ImplP1:SystemP1}@*)(((MappedSystem [| {| RPerrorAck |} |] Assumption_RPerrorAck)(*@\label{csp:ImplP1:SystemP1:hide-RPerrorAck}@*)\{|RPerrorAck|})
               [| {| RPpow24V |} |]
               Assumption_RPpow24V)\{| RPpow24V |}(*@\label{csp:ImplP1:SystemP1:end}@*)

 -- Subsequent composition with process RPEventMapping that relates the event RPsetPoint
 -- with an interrupt-driven version, suitable for comparison with SpecP1.
 RPSystemP1 = (SystemP1[[ RPsetPoint <- int_RPsetPoint ]](*@\label{csp:ImplP1:RPSystemP1}@*)
               [| {|int_RPsetPoint|} |]
               RPEventMapping(0))\{|int_RPsetPoint|}
 
 -- RPEventMapping holds the current value of the setpoint, as set via RPsetPoint, and that
 -- can be obtained via int_RPsetPoint.
 RPEventMapping(x) = RPsetPoint?nv -> RPEventMapping(nv)
                     []
                     int_RPsetPoint!x -> RPEventMapping(x)

 -- Process that synchronises on RPsetPoint and RPActualHV_out to offer their absolute
 -- difference via channel e.
 Error(actualhv,setpoint) = RPsetPoint?x -> Error(actualhv,x)
                            []
                            RPActualHV_out?x -> Error(x,setpoint)
                            []
                            e!abs_diff(actualhv,setpoint)(*@\label{csp:ImplP1:Error:e-abs-diff}@*) -> Error(actualhv,setpoint)(*@\label{csp:ImplP1:Error:end}@*)

 -- Process that every time unit requires immediate synchronisation on RPActualHV_out 
 -- followed by e.
 Sampler = EndBy(RPActualHV_out?x -> e?x -> SKIP,0); WAIT(1); Sampler
 
 -- Composition of RPSystemP1 and Error and Sampler, that relate the channels: 
 -- RPsetPoint, RPActualHV_out and e.
 ESystemP1 = (RPSystemP1(*@\label{csp:ImplP1:ESystemP1}@*)
              [| {|RPsetPoint,RPActualHV_out|} |] 
              (Error(0,0) [| {|RPActualHV_out, e|} |] Sampler)
             )\{|RPActualHV_out|}(*@\label{csp:ImplP1:ESystemP1:end}@*)
 
 -- Composition of ESystemP1 with the model of Assumption 3.
 ImplP1 = timed_priority(ESystemP1 [| {|RPsetPoint|} |] Assumption_SetPoint)(*@\label{csp:ImplP1:ImplP1}@*)
}
/@csp-end@/
\end{lstlisting}%
\vspace{-0.5ex}\caption{\label{csp:ImplP1}System interface for verification of \lstinline{SpecP1}.}\vspace{-1.25ex}
\end{lstfloat}

Moreover, we also explicitly capture three assumptions, that are implicitly required for the verification of \textbf{P1}: (1) the 24V power is stable, as reported via the input \RC{RPpow24V} (2) no error is to be acknowledged via \RC{RPerrorAck} (3) the HVC control software is correctly initialised, that is, \RC{RPsetPoint} has a value of zero during the first two cycles of \RC{State\_Machine}, so as not to trigger an error, and that the value of \RC{RPsetPoint} changes no more often than once per second. It should be noted that this third assumption regarding the frequency of change of \RC{RPsetPoint}, is more conservative than necessary, as \RC{RPsetPoint} is known to never change faster than within 10 seconds from the previous change, as mentioned in~\cref{sec:HVC}. These assumptions together define the normal working behaviour of the HVC, where the \RC{State\_machine} operates within the \RC{ClosedLoop} state, during which \textbf{P1} is required to hold.

Process \lstinline{ImplP1} is defined on \cref{csp:ImplP1:ImplP1} of~\cref{csp:ImplP1} as the parallel composition of \lstinline{ESystemP1}, synchronising on event \lstinline{RPsetPoint} with \lstinline{Assumption_SetPoint} (\cref{csp:ImplP1:Assumption_SetPoint}). The latter captures the first assumption by requiring that initially the setpoint is set to zero, with immediate effect, via the use of the \lstinline{EndBy} construct of $tock$-CSP, and where, after 22 milliseconds, its value can change arbitrarily, at most once per second, as defined by the process \lstinline{RPChange} on \cref{csp:ImplP1:RPChange}. Here 22~ms corresponds to at least two cycles of the execution of \RC{State\_machine}, given that for verification we consider each time unit as encoding 2ms, and that \RC{cycleTime} is instantiated as 10ms. The process \lstinline{ESystemP1} introduces the event \lstinline{e} in the context of the system behaviour, as defined by \lstinline{RPSystemP1} on \cref{csp:ImplP1:RPSystemP1}, that captures the other two assumptions and relates the \lstinline{RPsetPoint} of~\cref{fig:HVC-framework}, a sampled input, with the \lstinline{RPsetPoint} of \lstinline{SpecP1}, which is used as an interrupt for the purpose of specification.

\lstinline{ESystemP1} is defined on \crefrange{csp:ImplP1:ESystemP1}{csp:ImplP1:ESystemP1:end} as the parallel composition of \lstinline{RPSystemP1} and two processes \lstinline{Error} and \lstinline{Sampler}, that are also composed in parallel, synchronising on \lstinline{RPsetPoint} and \lstinline{RPActualHV_out}. The process \lstinline{Error} synchronises on these events so that it offers to synchronise on event \lstinline{e} with a value given by the absolute difference, specified by the application of \lstinline{abs_diff} on \cref{csp:ImplP1:Error:e-abs-diff}. This follows the definition of Property~\textbf{P1} as presented in~\cref{sec:properties}. The process \lstinline{Sampler} ensures that the actual high-voltage, read via \lstinline{RPActualHV_out}, and the difference, via \lstinline{e}, are updated exactly every time unit. This is specified by imposing a deadline of zero time units on the prefixing of \lstinline{RPActualHV_out} and \lstinline{e}, using a deadline, followed by a delay of exactly one time unit. This is a modelling mechanism to ensure that the events corresponding to sampled inputs or outputs, namely \lstinline{RPActualHV_out}, are updated regularly without introducing erroneous Zeno behaviours, that is, to prevent the CSP model from making an infinite number of updates within a finite amount of time.

The process \lstinline{RPSystemP1} is defined as a parallel composition (\cref{csp:ImplP1:RPSystemP1}) of the process \lstinline{SystemP1}, where the event \lstinline{RPsetPoint} is renamed to a new event \lstinline{int_RPsetPoint}, used in the synchronisation set, with \lstinline{RPEventMapping(0)}. The latter process takes in new values via \lstinline{RPsetPoint}, and then offers to synchronise on \lstinline{int_RPsetPoint} with the same value. The hiding of event \lstinline{int_RPsetPoint} makes it possible for \lstinline{SystemP1} to query the setpoint value periodically via \lstinline{int_RPsetPoint}, rather than directly via \lstinline{RPsetPoint}, as required for the comparison with \lstinline{SpecP1}. This is a modelling mechanism to ensure the event \lstinline{RPsetPoint} can be treated in the interrupt style of \lstinline{SpecP1}.

Finally, \lstinline{SystemP1}, defined on \crefrange{csp:ImplP1:SystemP1}{csp:ImplP1:SystemP1:end}, is the composition of the behaviour established by the co-verification framework, that accounts for the software and hardware modelling, as defined by \lstinline{MappedSystem} and processes \lstinline{Assumption_RPerrorAck} and \lstinline{Assumption_RPpow24V} that capture the second and third assumption for the purpose of verifying Property~\textbf{P1}. Here \lstinline{Assumption_RPpow24V} (\cref{csp:ImplP1:Assumption_RPpow24V}) initially sets the input \lstinline{RPpow24V} to the value 24, while \lstinline{Assumption_RPerrorAck} (\cref{csp:ImplP1:Assumption_RPerrorAck}) refuses to acknowledge any error via \lstinline{RPerrorAck} by behaving as \lstinline{STOP}, the process that deadlocks. As before, the use of hiding, on lines \ref{csp:ImplP1:SystemP1:hide-RPerrorAck} and \ref{csp:ImplP1:SystemP1:end}, completes the abstraction as the events \lstinline{RPerrorAck} and \lstinline{RPpow24V} are not relevant for refinement checking of \lstinline{SpecP1}. Next, we report on the use of FDR for checking the \lstinline{timed assertion P1}.

\begin{lstfloat}[!t]
\begin{lstlisting}
timed csp Instantiations /@csp-begin@/
nametype core_nat = { 0..1}
nametype core_real = { 0..2}
nametype core_int = { 0..1}
nametype core_boolean = Bool
nametype duty = { 0..100}

overLimitF(x) = if x > 2 then 2 else x
underLimitF(x) = if x < 0 then 0 else x

ms(t) = t1/2
s(t) = t1*1000/2
// ...
/@csp-end@/
\end{lstlisting}
\vspace{-0.5ex}\caption{\label{csp:Instantiations}Instantiation of data-types as finite sets, and definition of functions used in the RoboChart model, required for model-checking with FDR.}\vspace{-0.5ex}
\end{lstfloat}

\paragraph{Verification Parameters and Results}\label{par:verification_results}

For model-checking with FDR, not only constants of the RoboChart model have to be instantiated, but the domain of the data-types must also be defined as discrete finite sets. These are defined in a special \lstinline{csp} block named \lstinline{Instantiations}, reproduced in~\cref{csp:Instantiations}.

Besides, in this block we also give a \texttt{CSP$_\text{\texttt{M}}$} definition for all of the functions declared in the RoboChart model. \RC{overLimitF} and \RC{underLimitF}, used by the software operation \RC{AdjustLimits}, ensure that the result is closed under the type. Since in the software model all time units are divisible by 2, the smallest time unit is chosen as encoding 2 milliseconds, thus the function \RC{ms}, halves the argument, and \RC{s}, encoding seconds, is defined analogously. 

The reals are instantiated as the set $\{0,1,2\}$ as this is a realistic representation of the different inputs and outputs, namely \RC{RPsetPoint}, where values from 0, 1 and 2 naturally map to high-voltage values 0, 40 and 80~kV. Recall that, from a paint robot application point of view, it is given that once the high-voltage is activated and turned on, it requires values larger than $30 kV$, i.e., $HV\_SetPoint \in 0 \cup [30 \: 90],  kV$. Hence by switching between the values 0, 40 and 80~kV, we are able to capture all possible qualitative combinations for HV setpoint changes (i.e., charge/discharge as well as increase/decrease in setpoint). Based on this, it is observed that the set $\{0,1,2\}$ provides a rich enough representation of the system inputs.

The \lstinline{timed assertion P1} is successfully verified by FDR. On a dual AMD EPYC 7501 32-core machine with 1TiB of RAM, it took FDR 2850s overall to verify that the property holds (1456s to compile the Labelled Transition System (LTS), and 1394s to verify the refinement), having visited 126,481,225 states and 517,333,656 transitions. For comparison, in~\cref{tab:softare-verification}, we include this result together with those concerning only the verification of software properties, which we address next.

\section{Formal Verification of Software Properties}
\label{sec:cover:software-properties}
In what follows we discuss the verification of properties \textbf{P2}-\textbf{P4} of~\cref{sec:properties}, which concern only the software. Property~\textbf{P4} concerning deadlock freedom can be specified directly using the assertion language provided by RoboTool. Properties \textbf{P2} and \textbf{P3}, on the other hand, are specified directly in \texttt{CSP$_\text{\texttt{M}}$}.

\begin{lstfloat}[!th]
\begin{lstlisting}
timed csp SpecP2 /@csp-begin@/
Timed(OneStep) {
 SpecP2 = timed_priority(PWM_off)

 -- A duty cycle value can be received via int_dutyCyclePWM1.
 PWM_Behaviour = mod_sys::int_dutyCyclePWM1.out?x:{x | x <- duty, x > 0} -> PWM_on
                 []
                 mod_sys::int_dutyCyclePWM1.out.0 -> PWM_off

 -- If the duty cycle is 0, then the PWM can also be turned off.
 PWM_off = (*@\label{csp:SpecP2:PWM_off}@*)PWM_Behaviour [] mod_sys::ext_pow24VStatus.in.Power_Off -> PWM_off

 -- If the duty cycle is currently greater than 0, then if the ext_pow24VStatus goes Off,
 -- so should the PWM within 10ms.
 PWM_on = (*@\label{csp:SpecP2:PWM_on}@*)
  PWM_Behaviour
  []
  mod_sys::ext_pow24VStatus.in.Power_Off ->
   ADeadline({|mod_sys::ext_pow24VStatus.in.Power_Off,mod_sys::int_dutyCyclePWM1.out|},
             {|mod_sys::int_dutyCyclePWM1.out.0|},ms(10)) ; PWM_off(*@\label{csp:SpecP2:PWM_on:end}@*)
}
/@csp-end@/

timed csp mod_sys_pwm /@associated to@/ mod_sys /@csp-begin@/
Timed(OneStep) {
 mod_sys_pwm = (*@\label{csp:SpecP2:mod_sys_pwm}@*)
  timed_priority(
   mod_sys::O__(0,ms(10),1) 
   |\ {|mod_sys::ext_pow24VStatus.in.Power_Off,mod_sys::int_dutyCyclePWM1.out,tock|})(*@\label{csp:SpecP2:mod_sys_pwm:end}@*)
}
/@csp-end@/

(*@\label{csp:SpecP2:P2}@*)timed assertion P2 : mod_sys_pwm refines SpecP2 /@in the traces model@/ // Actual check for P2.
\end{lstlisting}
\caption{\label{csp:SpecP2}Specification for Property \textbf{P2} and \lstinline{assertion P2}.}
\end{lstfloat}

\paragraph{Property \textbf{P2}} Taking into account the RoboChart model, \textbf{P2} can be restated as requiring that the observation of the input \RC{ext\_pow24VStatus} with value \RC{Power::Off} is followed by the output \RC{int\_dutyCyclePWM1} with value \RC{0}. As CSP adopts a reactive paradigm, the process \lstinline{SpecP2}, specified in~\cref{csp:SpecP2}, is defined in terms of events. It considers the case when the output \RC{int\_dutyCyclePWM1} has been set to a value other than zero and subsequently \RC{ext\_pow24VStatus} is observed with value \RC{Power::Off}.

The behaviour of \lstinline{SpecP2} is that of \lstinline{PWM_off}, defined on \cref{csp:SpecP2:PWM_off} as an external choice over behaving as \lstinline{PWM_Behaviour} or accepting the event \lstinline{mod_sys::ext_pow24VStatus.in.Power_Off}, followed by the recursion on \lstinline{PWM_off}. 
\lstinline{PWM_Behaviour} tracks the changes of the output \RC{int\_dutyCyclePWM1} by offering a value greater than 0 and then behaving as \lstinline{PWM_on}, or, a value of 0, and then behaving as \lstinline{PWM_off}. In \lstinline{PWM_on} (\crefrange{csp:SpecP2:PWM_on}{csp:SpecP2:PWM_on:end}) we capture the core of Property~\textbf{P2}, where, following the event \lstinline{mod_sys::ext_pow24VStatus.in.Power_Off} we require \lstinline{mod_sys::int_dutyCyclePWM1.out.0} to be observed within 10 milliseconds (matching the \RC{cycleTime} used by \RC{State\_machine}) using the process \lstinline{ADeadline}, after which the process behaves as \lstinline{PWM_off} again as specified by the sequential composition on \cref{csp:SpecP2:PWM_on:end}.

The assertion for verifying Property~\textbf{P2} is written as the \lstinline{timed assertion P2}, on \cref{csp:SpecP2:P2} of~\cref{csp:SpecP2}.
It is stated as a refinement assertion \lstinline{/@in the traces model@/}. The process \lstinline{mod_sys_pwm} is defined on \crefrange{csp:SpecP2:mod_sys_pwm}{csp:SpecP2:mod_sys_pwm:end} by constraining the timed semantics of \RC{mod\_sys} and hiding every \CSPM~event other than those mentioned by \lstinline{SpecP2} (including the implicit $tock$) using the projection operator ({\lstinline!|\!}) so that the comparison is meaningful.

\begin{lstfloat}[th]
\begin{lstlisting}
timed csp SpecP3 /@csp-begin@/
Timed(OneStep) {
 SpecP3 = timed_priority(mSetPoint_zero)
 mSetPoint = mod_sys::set_mSetPoint?x:{x | x <- core_real, x > 0} -> mSetPoint_non_zero
	            []
	            mod_sys::set_mSetPoint.0 -> mSetPoint_zero

 mSetPoint_zero = mSetPoint [] mod_sys::ext_pow24VStatus.in.Power_Off -> mSetPoint_zero
	  	  
 mSetPoint_non_zero = 
  mSetPoint_zero
  []
  mod_sys::ext_pow24VStatus.in.Power_Off ->
   ADeadline({|mod_sys::ext_pow24VStatus.in.Power_Off,mod_sys::set_mSetPoint|},
             {|mod_sys::set_mSetPoint.0|},ms(10)) ; mSetPoint_zero
}
/@csp-end@/

// Constrained form of mod_sys for P3
timed csp mod_sys_setpoint /@associated to@/ mod_sys /@csp-begin@/ 
Timed(OneStep) {
 mod_sys_setpoint = 
  timed_priority(
   mod_sys::AS_O__(0,ms(10),1) 
   |\ {|mod_sys::ext_pow24VStatus.in.Power_Off,mod_sys::set_mSetPoint,tock|})
}
/@csp-end@/

// Actual check for P3
timed assertion P3 : mod_sys_setpoint refines SpecP3 /@in the traces model@/
\end{lstlisting}
\caption{\label{csp:SpecP3}Specification for Property \textbf{P3}.}
\end{lstfloat}

\paragraph{Property \textbf{P3}} The next property, \textbf{P3}, is specified in \CSPM~by the process \lstinline{SpecP3}, defined in~\cref{csp:SpecP3}.
The structure is similar to \lstinline{SpecP2}, and it also uses the event \lstinline{ext_pow24VStatus}. Unlike~\lstinline{SpecP2}, however, the process \lstinline{SpecP3} tracks changes in the assignment of values to the shared variable \RC{mSetPoint}, encoded in the RoboChart semantics via events \lstinline{set_mSetPoint}. We observe that since \RC{mSetPoint} is a variable of the software, rather than an output of \RC{mod\_sys}, such an assignment is not visible in the semantics of a RoboChart module. Instead, we use a tailored version of the semantics, calculated by RoboTool, that supports this type of analysis, in a similar way to how state reachability checks are implemented. The actual check for Property \textbf{P3} is specified by \lstinline{assertion P3}, a refinement that considers the process \lstinline{mod_sys_setpoint}, a constrained form of \lstinline{mod_sys}, defined similarly to process \lstinline{mod_sys_pwm} in \lstinline{assertion P2}.

\begin{lstfloat}[th]
\begin{lstlisting}
//P4: Checks if the model is deadlock free
(*@\label{csp:deadlock-free-states:P4}@*)timed assertion P4 : mod_sys is /@deadlock-free@/

// State reachability checks
(*@\label{csp:deadlock-free-states:reach-begin}@*)timed assertion Reach_Init          : State_machine::Init /@is reachable in@/ mod_sys
timed assertion Reach_Wait24VPower  : State_machine::Wait24Vpower /@is reachable in@/ mod_sys
timed assertion Reach_ClosedLoop    : State_machine::ClosedLoop /@is reachable in@/ mod_sys
timed assertion Reach_ErrorMode     : State_machine::ErrorMode /@is reachable in@/ mod_sys
timed assertion Reach_Watchdog_s0   : Watchdog::s0 /@is reachable in@/ mod_sys
(*@\label{csp:deadlock-free-states:reach-end}@*)timed assertion Reach_Watchdog_s1   : Watchdog::s1 /@is reachable in@/ mod_sys
\end{lstlisting}
\caption{\label{csp:deadlock-free-states}Deadlock freedom and state reachability checks.}
\end{lstfloat}

\paragraph{Property \textbf{P4}} The fourth property requires that the software is deadlock free. This is directly specified using the RoboChart assertion \lstinline{/@is deadlock-free@/} on \cref{csp:deadlock-free-states:P4} of~\cref{csp:deadlock-free-states}.

A timed deadlock manifests when the system refuses to perform any event, but time may pass indefinitely. Its absence is checked in FDR in the failures-divergences semantic model of CSP, using a technique inspired by Roscoe~\cite{Roscoe2013a}, that effectively checks that no state configuration is reached whereby an infinite amount of time can pass while refusing to perform every regular event.

Moreover, as a sanity check, we also verify that all of the states of \RC{State\_Machine} and \RC{Watchdog} are reachable, using the \lstinline{assertion}s on lines \ref{csp:deadlock-free-states:reach-begin} to \ref{csp:deadlock-free-states:reach-end} of~\cref{csp:deadlock-free-states}.

Similarly to the verification of Property~\textbf{P3}, for checking reachability, RoboTool uses a tailored version of the semantics whereby the entrance of states is visible, as detailed in~\cite{RoboChartSoSym}.

\begin{table}[tp]
\centering
\begin{tabular}{|l|c|l|l|l|l|l|}
\hline
\multicolumn{1}{|c|}{\multirow{2}{*}{\textbf{Assertion}}} 
& \multicolumn{1}{c|}{\multirow{2}{*}{\textbf{Result}}} 
& \multicolumn{3}{c|}{\textbf{Elapsed Time}} 
& \multicolumn{2}{c|}{\textbf{Complexity}} 
\\ \cline{3-7} 
\multicolumn{1}{|c|}{}
& \multicolumn{1}{c|}{}
& \multicolumn{1}{c|}{\textbf{Compilation}}
& \multicolumn{1}{c|}{\textbf{Verification}}
& \multicolumn{1}{c|}{\textbf{Total}}
& \multicolumn{1}{c|}{\textbf{States}} 
& \multicolumn{1}{c|}{\textbf{Transitions}} 
\\ \hline
{\lstinline!P1!}                    & $\checkmark$  & 1456s & 1394s & 2850s     & 126,481,225 & 517,333,656         \\ \hline
{\lstinline!P2!}                    & $\checkmark$  & 1456s & 247s  & 1703s     & 1,460,749   & 3,855,659           \\ \hline
{\lstinline!P3!}                    & $\checkmark$  & 1539s & 248s  & 1787s     & 1,452,829   & 3,831,246           \\ \hline
{\lstinline!P4!}                    & $\checkmark$  & 1253s & 334s  & 1587s     & 1,920,070   & 5,795,521           \\ \hline
{\lstinline!Reach_Init!}            & $\checkmark$  & 789s  & 1.07s & 790.07s   & 3,292       & 12,455              \\ \hline
{\lstinline!Reach_Wait24VPower!}    & $\checkmark$  & 789s  & 5.51s & 794.51s   & 2,229,843   & 9,672,801           \\ \hline
{\lstinline!Reach_ClosedLoop!}      & $\checkmark$  & 789s  & 11.62s& 800.62s   & 8,148,391   & 35,349,260          \\ \hline
{\lstinline!Reach_ErrorMode!}       & $\checkmark$  & 789s  & 10.38s& 799.38s   & 6,756,722   & 29,260,634          \\ \hline
{\lstinline!Reach_Watchdog_s0!}     & $\checkmark$  & 789s  & 0.60s & 789.06s   & 352         & 976                 \\ \hline
{\lstinline!Reach_Watchdog_s1!}     & $\checkmark$  & 789s  & 0.80s & 789.08s   & 1,420       & 4,667               \\ \hline
\end{tabular}
\cprotect\caption{Results of model-checking Properties~\textbf{P1}-\textbf{P4}, as well as reachability analysis using FDR.}
\label{tab:softare-verification}
\end{table}

\paragraph{Verification Results}

The results of model-checking are summarised in~\cref{tab:softare-verification}. The time elapsed is the sum of the time taken to compile and verify the Labelled Transition System (LTS), as calculated by FDR, on a dual AMD EPYC 7501 32-core machine with 1TiB of RAM. Complexity is broken down into number of states and transitions visited when verifying the assertions. Compilation takes longer than verification as the \CSPM~automatically generated by RoboTool employs compression functions to minimize the LTS. The compression algorithms used by FDR are largely sequential, whereas verification can exploit multiple cores efficiently. Verification of~\textbf{P1} is more complex than the verification of software-only properties, due to the mechanisation of both the framework and the hardware abstraction.

\section{Concluding Remarks and Future Work}
\label{sec:conclusion}
Co-simulation, e.g., effectively combining various types of models and simulation tools in order to reach system-level results, is a rather well-known and established industrial practice that has received recent attention~\cite{INTOCPS}. This paper advocates extension of the same school of thought and practice into the formal verification domain. Centering the focus around the paint robot HVC application, this paper guides the reader through an industrial use-case of co-verification where modelling and verification results from different tools are lifted into a unifying framework, thereby allowing the verification of system-level properties. 

In our case study, we have used RoboChart for modelling the software, and Simulink for hardware modelling. RoboChart models are typically of a higher abstraction level than those used for dynamic simulation. Therefore, abstractly capturing the behaviour of low-level software, like that of the HVC, can be challenging, especially for practitioners who are more familiar with dynamic simulation. Another aspect of practical concern is finding the right level of abstraction to achieve computationally tractable results for model-checking. Simulink, on the other hand, is convenient for modelling and simulation of dynamics, but is limited in the ability to perform verification. Continuous blocks need to be discretized for use with SDV, and, on a more practical level, it is not always clear whether counter-example generation is feasible.

Because of the general form of our approach, we envision that the principle of connecting various hardware and software components, coupled via platform mappings, also advocated in~\cite{diagrammatic_RoboSim_models}, could constitute a useful, and pragmatic, basis for use with other formal verification tools and techniques. An avenue for future work could include the complete formalisation and mechanisation of decomposition patterns using a unifying semantic framework like the UTP~\cite{HH98}, that caters for multiple paradigms, with support for interactive theorem proving available via Isabelle/UTP~\cite{Foster2020}.

Another possibility for generalization is based on the observation that the HVC application considered in this paper has a rather generic form of a feedback control system and as such, has  rather natural and generic properties that are expected to be fulfilled. Hence, the co-verification framework and experiences reported on \changed[\C{2}]{here} are also highly relevant for any cyber-physical system tracking a setpoint reference.

On a more practical level, we anticipate that the automatic generation of proof models from a declarative notation capturing the framework outlined in~\cref{fig:HVC-framework}, namely for model-checking as we do in our case study, could be useful for practitioners. Such work could also address the dichotomy between the use of events to represent sampling of inputs, and their use as interrupts in the style of CSP, that facilitates the specification of properties like \textbf{P1}. Not surprisingly, this is a paradigm shift also seen in the conformance relation between RoboChart and the closely-related simulation language RoboSim~\cite{CavalcantiSMRFD19}.

The results presented in this paper, extend those provided in~\cite{FV_of_HVC_SBMF_2020} by performing verification against an older version of the HVC software, which was known to be faulty, \emph{e.g.}, had been observed in practice to generate solutions contradicting some of the properties listed in~\cref{sec:properties}. Our new results revealed that neither Property~\textbf{P2} nor~\textbf{P3} were satisfied. This gives credence to our ability to successfully capture the critical behavioural aspects of the HVC. Moreover, as these errors went undetected by traditional testing methods in an earlier version of the software, the results in this paper also serve as a testimony of the strength and suitability of using formal verification methods for industrial safety critical systems. The interested reader is referred to consult~\cite{FV_of_HVC_SBMF_2020} for more detailed information about these previous errors.

Related to our motivation in addressing the ongoing industrial trends in robotics, where an increasing number of safety features and functions are handled by software, we also acknowledge the emergence of adaptive or learning-based software components. The use of machine learning methods, and their inherent opaqueness, presents significant challenges in fulfilling certification requirements and obtaining wide-scale market acceptance. To push forward the socio-technical research frontier drastically and improve current practices of robotic system design, analysis and verification is scope of future research.

Related work is ongoing, for example, to provide facilities in RoboChart to capture properties of neural networks at a suitable level of abstraction, which could be a useful basis in the future to extend our co-verification approach. In addition, for RoboChart models featuring probabilistic junctions~\cite{WoodcockC00Y19,Ye21}, RoboTool also supports the generation of reactive modules for analysis with PRISM~\cite{PRISM}, which features both probabilistic and statistical model-checking. In the future, we plan to explore this avenue for verification, namely for reasoning in the presence of faults and uncertainty.

\section*{Acknowledgements}

The authors would like to gratefully acknowledge all the support, guidance and inspiration provided by Prof.\ Ana Cavalcanti during this work. 
The research presented in this paper has received funding from the Norwegian Research Council, SFI Offshore Mechatronics, project number 237896. Pedro Ribeiro is funded by the UK EPSRC (grant EP/M025756/1), and by the Royal Academy of Engineering (grant CiET1718/45). The icons used in RoboChart have been made by Sarfraz Shoukat, Freepik, Google, Icomoon and Madebyoliver from \url{www.flaticon.com}, and are licensed under CC 3.0 BY.

\pagebreak
\bibliographystyle{elsarticle-num}
\changed[\C{3}]{
\bibliography{HVC.bib}
}

%%%%%%%%%%%%%%%%%%%%%%%%%%%%%%%%%%%%%%%%%%%%%%%%%%%%%%%%%%%%%%%%%%%
%%%%%%%%%%%%%%%%%%%%%%%%%%%%%%%%%%%%%%%%%%%%%%%%%%%%%%%%%%%%%%%%%%%
\ifdefined\CHANGES \indexprologue{This index lists for each comment the pages
where the text has been modified to address the comment. Since the same page
may contain multiple changes, the page number contains the index of the
change in superscript to identify different changes. Finally, the page number
contains a hyperlink that takes the reader to the corresponding change.}
\printindex[changes] \fi
\end{document}